\documentclass{article}
\usepackage{spconf,amsmath,amssymb,graphicx,hyperref}
\usepackage{booktabs}
\usepackage{multirow}
\usepackage{array}
\usepackage{float}
\usepackage{tipa}
\hypersetup{hidelinks}

\title{SIMULATE, RECORD, VERIFY: A LANGUAGE-PORTABLE FRAMEWORK FOR MUSCLE-GROUNDED ARTICULATORY QA}
\name{
Seungho Eum$^{1}$ \qquad
Shantong Sun$^{3}$ \qquad
Unsang Park\sthanks{Corresponding author: unsangpark@sogang.ac.kr}$^{1,2}$
}
\address{
$^{1}$ Department of Computer Science and Engineering, Sogang University, Seoul, Korea \\
$^{2}$ Department of Artificial Intelligence, Sogang University, Seoul, Korea\\
$^{3}$ School of Computer Science and Communication Engineering, Jiangsu University, ZhenJiang, China}

\makeatletter
\let\sp@thebibliography\thebibliography
\def\thebibliography#1{\sp@thebibliography{#1}%
  \small
  \setlength{\itemsep}{0pt}\setlength{\parsep}{0pt}\setlength{\topsep}{0pt}}
\makeatother

\begin{document}
%\ninept
%
\maketitle
\begin{abstract}

Articulatory data describe what the tongue looks like, but geometry alone does not explain the muscle-level why behind a configuration or how to move it toward a target posture. We present a simulator-based framework that constructs this supervision from controlled biomechanical inputs. Each simulated configuration is stored with its generating muscle state and geometric properties in a structured fact record. Gold answers are derived before linguistic realization, and every naturalized output is checked against its source record, so the same records support new languages and question types without re-simulation.

We instantiate the framework as 3DTongueQA using the ArtiSynth Badin tongue model. From 295,115 meshes, we construct 891,156 record-checked QA instances per language, and the checker detects over 96.9\% of injected corruptions. Korean and Spanish realizations and a new question type are added from the same records, confirming portability. A SpiralNet++–Qwen3 probe reaches 62.9 Muscle EM, above 44.0 for nearest-neighbor transfer and 7.2 for a zero-shot commercial LLM given the same mesh, while mesh shuffling drops it to near zero. The supervision is learnable, useful beyond retrieval, and grounded in paired geometry. It reflects simulator-defined states rather than measured physiology. Code, templates, and QA sets are available at https://github.com/esh0504/muscle-grounded-qa.

\end{abstract}

\vspace{-3.5mm}
\section{Introduction}
\label{sec:intro}
\vspace{-2mm}

Visual feedback of tongue movement can support pronunciation learning \cite{katz2015visual}. Existing articulatory data can show \emph{what} the tongue looks like, but geometry alone does not explain \emph{why} a configuration arises or \emph{how} it should change toward a target posture. Such explanations require supervision that links tongue geometry with the muscle state that generated it and with the muscle adjustments needed to approach a target.

Real-time MRI and electromagnetic articulography capture articulatory motion \cite{narayanan2011multimodal,narayanan2014real} as a midsagittal slice or a few flesh points, but neither observes the muscle activations that generate each tongue configuration.

Biomechanical tongue models have long related tongue posture to muscle activation \cite{buchaillard2009biomechanical,stavness2012biomechanical,harandi2017variability}, and inverse modeling estimates activations from target geometry \cite{stavness2012automatic,tolpadi2018inverse}. However, these methods do not construct natural-language supervision that jointly describes generating states and target-directed adjustments. Existing 3D QA datasets also focus on objects and spatial relations \cite{ma2022sqa3d,azuma2022scanqa}, rather than explaining biomechanical shapes through their generating states and corrective actions. Meanwhile, synthetic QA with programmatic ground truth \cite{johnson2017clevr} and 3D-LLMs \cite{hong20233dllm,xu2024pointllm} show that structured facts and 3D observations can be used for language supervision.

We address this gap with a construction framework for muscle-grounded QA supervision that captures both \emph{why} a tongue configuration arises and \emph{how} it should change toward a target. Controlled muscle activations drive a biomechanical simulator, and each valid configuration is stored with its generating input as a structured fact record. Questions and gold answers are derived from these records before linguistic realization, which changes only the surface form and is checked against the source record. The same records are therefore reused across languages and question types while preserving provenance and verifiability.

We instantiate the framework as 3DTongueQA using the ArtiSynth Badin finite-element tongue model \cite{lloyd2012artisynth,vogt2006efficient,badin1998three}. Our evaluation follows the construction pipeline. Sec.~\ref{sec:res-verif} evaluates record-based verification, Sec.~\ref{sec:res-port} reuse across languages and question types, and Sec.~\ref{sec:res-utility} learnability and geometry grounding. The final evaluation uses a SpiralNet++~\cite{gong2019spiralnet++}--Qwen3-8B~\cite{yang2025qwen3} probe against the controls in Table~\ref{tab:results}, with implementation details in \cite{extended3dtongueqa}.

\noindent\textbf{Contributions.}
We introduce a provenance-preserving framework that constructs supervision for both the simulator-defined generating muscle state behind an observed tongue configuration and the muscle adjustments needed to approach a target posture. We separate factual construction from linguistic realization so that the same records serve multiple languages and question types without re-simulation. We also introduce 3DTongueQA and show that its supervision is learnable, geometry-grounded, and usable through both a language interface and task-specific readouts.

\begin{table*}[t]
\centering
\caption{
Probe results on 400 anchor-balanced sample-held-out meshes with
1,200 automatic items and 30 open-ended prompts per English
LLM-based model. Automatic metrics use a 0--100 scale and human
ratings 1--5.
``Shuf.'' denotes mesh shuffling and ``--'' unreported scores.
Bold marks the highest English score per column, not
significance, and $\dagger$ marks the readout that also receives
the target activation vector.
}
\label{tab:results}

\begingroup
\fontsize{9}{10}\selectfont
\setlength{\tabcolsep}{2.0pt}
\renewcommand{\arraystretch}{1.0}

\begin{tabular}{@{}lcccccccc@{}}
\toprule
& \multicolumn{2}{c}{Muscle EM}
& \multicolumn{2}{c}{Geom. Value Acc.}
& \multicolumn{2}{c}{Direction EM}
& \multirow{2}{*}{Fluency}
& \multirow{2}{*}{Factual Acc.} \\
\cmidrule(lr){2-3}
\cmidrule(lr){4-5}
\cmidrule(lr){6-7}
Method
& Paired & Shuf.
& Paired & Shuf.
& Paired & Shuf.
& & \\
\midrule

\multicolumn{9}{@{}l}{\emph{Reference baselines}} \\

GPT-5 Pro, zero-shot~\cite{GPT-pro}
& 7.2 & --
& 28.7 & --
& 35.8 & --
& \textbf{3.50} & 2.23 \\

Text-only LLM
& 0.0 & --
& 49.6 & --
& 39.4 & --
& 2.34 & 1.77 \\

1-NN answer transfer
& 44.0 & --
& \textbf{97.2} & --
& 47.3 & --
& -- & -- \\

\midrule

\multicolumn{9}{@{}l}{
\emph{English geometry-conditioned probes}
} \\

2D Muscle-Aware + LLM
& 42.1\,$\pm$\,3.9 & 2.2\,$\pm$\,0.3
& 50.9\,$\pm$\,1.3 & 30.0\,$\pm$\,0.7
& 55.9\,$\pm$\,1.1 & 38.5\,$\pm$\,0.8
& 2.86 & 2.33 \\

3D Mesh-AE + LLM
& 43.0\,$\pm$\,5.6 & 2.8\,$\pm$\,0.4
& 76.5\,$\pm$\,0.6 & 34.9\,$\pm$\,0.3
& 63.3\,$\pm$\,2.8 & 39.7\,$\pm$\,0.5
& 3.08 & 2.70 \\

3D Muscle-Aware + LLM
& 62.9\,$\pm$\,9.2 & 2.2\,$\pm$\,0.5
& 74.0\,$\pm$\,0.2 & 31.2\,$\pm$\,0.2
& 65.9\,$\pm$\,4.7 & 40.9\,$\pm$\,0.7
& 2.72 & \textbf{3.81} \\

Task-specific structured readouts
& \textbf{88.7\,$\pm$\,0.7} & 1.2\,$\pm$\,0.0
& 87.5\,$\pm$\,1.1 & 41.6\,$\pm$\,0.6
& \textbf{93.3\,$\pm$\,1.0}$^{\dagger}$ & 1.8\,$\pm$\,0.0
& -- & -- \\

\midrule

\multicolumn{9}{@{}l}{
\emph{Korean instantiation on the same records}
} \\

3D Muscle-Aware + LLM, Korean
& 67.0 & 2.8
& 70.3 & 30.8
& 58.8 & 36.8
& 2.86 & 3.79 \\

\bottomrule
\end{tabular}

\par\vspace{2pt}
\begin{minipage}{\textwidth}
\fontsize{9}{10.5}\selectfont
Prompt-level Friedman tests across the five English
LLM-based models give $\chi^2(4)=13.03$, $p=0.011$ for Fluency
and $\chi^2(4)=58.10$, $p<0.001$ for Factual Accuracy.
\end{minipage}

\endgroup
\vspace{-5mm}
\end{table*}

\vspace{-3mm}
\section{The Construction Framework}
\vspace{-2mm}
\label{sec:pipeline}

\begin{figure}[h]
\centering
\includegraphics[width=0.82\columnwidth]{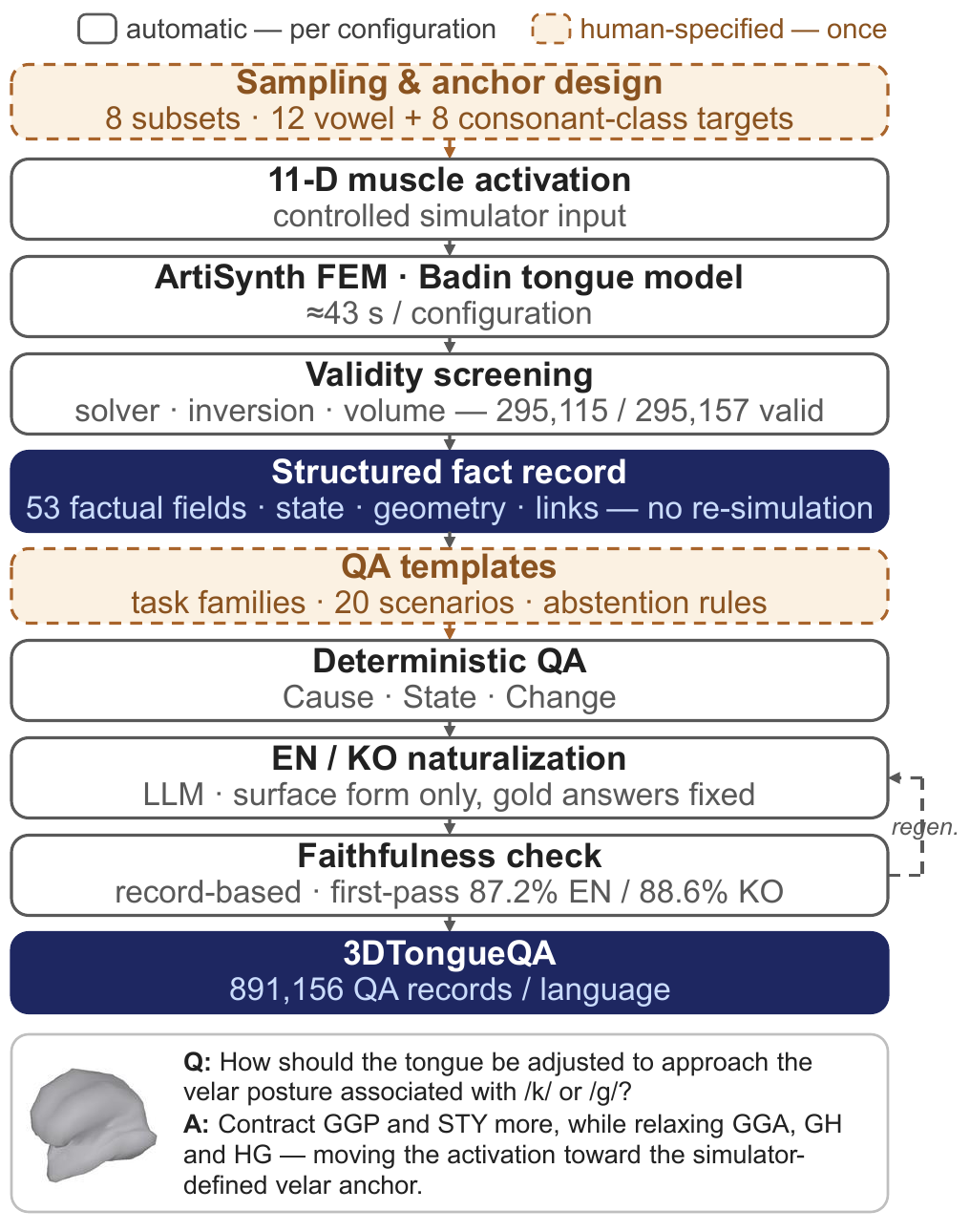}
\vspace{-5mm}
\caption{Construction framework with a target-directed example from 3DTongueQA. Dashed blocks are authored once, while the remaining stages run automatically for each configuration. Spanish follows the same language realization and verification pipeline.}
\label{fig:pipeline}
\end{figure}

The framework consists of three stages called simulate, record, and verify, as shown in Fig.~\ref{fig:pipeline}. Human authoring is required only for sampling and anchor design, QA templates with abstention rules, and a language-specific realization module with a verification lexicon. All remaining steps are automated.

\vspace{-4.5mm}
\subsection{Instantiation with Simulated Tongue Geometry}
\label{sec:simulation}

We use the ArtiSynth Badin finite-element tongue model
\cite{lloyd2012artisynth,vogt2006efficient,badin1998three}
to map 11-dimensional muscle-activation vectors to fixed-topology 3D tongue meshes. Sampling is divided into eight subsets covering isolated and interacting muscles, single-muscle interventions, dose--response trajectories, phoneme-associated postures, and broader feasible activation patterns.

To associate configurations with phonetic targets, we construct 12 vowel and 8 consonant-class anchors using literature-informed muscle patterns and articulatory constraints
\cite{buchaillard2009biomechanical,stavness2012biomechanical,takano2007mri}.
Gaussian perturbations around each anchor provide within-anchor variation, while interpolation between anchors provides transition postures. As an acoustic sanity check, model-derived $F_1$ and $F_2$ correlate with measured vowel formants at $r=0.98$ and $r=0.99$ across categories, preserving the relative vowel organization although the absolute $F_2$ range is compressed.

Given a sampling scheme, the framework automatically generates and screens large numbers of configurations. We reject solver failures, invalid elements, out-of-bound volume ratios, and unsettled runs. For the 3DTongueQA instantiation, we generate 295,157 configurations, of which 295,115 pass the screening stage.

\vspace{-2.5mm}
\subsection{Fact Records and Deterministic QA}
\label{sec:records}

For each valid configuration, we construct a structured fact record from the simulated mesh and its generating input. The record contains 53 factual fields describing muscle activations, constriction information, tongue-shape curvature, landmark coordinates, and other physical properties. It also stores links to anchors, interventions, and trajectories.

Deterministic generators use only these records to construct QA from 20 generation scenarios with abstention rules. The resulting QA covers three objectives. \emph{Cause} recovers the simulator-defined muscle state that generated an observed mesh. \emph{State} describes geometric and articulatory properties. \emph{Change} identifies the muscle adjustments needed to approach a target anchor. Undefined facts are skipped or assigned an explicit abstention. Every gold answer is therefore determined from the record before linguistic realization.

\vspace{-2.5mm}
\subsection{Language Realization and Record-Based Verification}
\label{sec:naturalization}

Template-based QA is factually controlled but linguistically limited, so we rewrite each templated question and answer with Qwen2.5-72B-Instruct \cite{hui2024qwen2}. This step changes only the linguistic surface form and never determines the gold answer.

Each language requires a realization module and a verification lexicon. The lexicon defines admissible surface forms for fields stored in the fact records. The checker normalizes language-specific expressions through the lexicon and compares the recovered facts with the source record. It verifies muscle identities, directions, numerical values, target phonemes, regions, and abstention decisions. Failed outputs are regenerated, while items that fail again retain their canonical rendering.

English and Korean are realized over the full corpus. To test language portability, we also add Spanish by translating the 20 scenario templates and constructing the corresponding verification lexicon.

The entire pipeline is released as code. Because every gold answer is a deterministic function of the 53 factual fields in a record, users can compose new question templates by combining these fields in any way that yields a determinate answer, and the released checker verifies the naturalized output against the same record. Adding a language requires only a realization module and a verification lexicon over the same fields, which took about 20 minutes of author time for Spanish and was checked through back-translation.
\vspace{-3.5mm}
\section{Evaluation Protocol}
\label{sec:evaluation}
\vspace{-2mm}

We evaluate three properties of the framework. \emph{Verifiability} measures whether generated QA remains consistent with its source records through automatic checking, corruption tests, and manual inspection. \emph{Portability and record reuse} evaluate whether the same records support new languages and question types without re-simulation. \emph{Utility} evaluates whether the resulting supervision is learnable and grounded in paired tongue geometry.

\noindent\textbf{Tasks and metrics.}
Utility is evaluated with three tasks derived from Cause, State, and Change. For each task, the LLM receives a fixed answer prefix and predicts the task-dependent continuation. \emph{Muscle} predicts the active-muscle set and is scored by exact match with the simulator-defined set. \emph{Value} predicts one of six mesh-derived scalar properties and is scored within feature-specific tolerances. \emph{Direction} predicts whether relevant muscle activations should increase, decrease, or remain unchanged toward a target posture and is scored by exact match of the muscle-direction set.

\noindent\textbf{Probe and test protocol.}
The unified QA probe combines a SpiralNet++ mesh encoder
\cite{gong2019spiralnet++}, a projection module, and Qwen3-8B
\cite{yang2025qwen3}. The muscle-aware encoder is pretrained on simulator activations and frozen during QA training, which uses 240k assistant turns (about 67k records, 7.5\% of the corpus) per language. Text-only removes geometry, 2D replaces the mesh with rendered views, and Mesh-AE replaces muscle-aware pretraining with geometry-oriented pretraining. Mesh shuffling breaks the question--mesh pairing, retrieval transfers answers from nearby training meshes, and task-specific readouts replace the LLM with fixed-schema heads. Zero-shot GPT-5 Pro receives the raw 3D mesh file together with the question, as a reference for general-purpose knowledge rather than a matched multimodal baseline.

We evaluate 400 sample-held-out meshes balanced across 20 anchor categories. Each mesh has one Muscle, Value, and Direction question, yielding 1,200 automatically scored items. English geometry-conditioned models report three-seed results, while Korean uses the same meshes and facts with one seed.

\noindent\textbf{Open-ended evaluation.}
Three model-blind speech researchers rate responses to 30 held-out English prompts per LLM-based model for Fluency and reference-based Factual Accuracy on 1--5 scales. The corresponding ICC values are 0.82 and 0.83. Korean uses a separate prompt set.

\vspace{-4.5mm}
\section{Results}
\label{sec:results}

\vspace{-2mm}
\subsection{Record-Based Verifiability}
\label{sec:res-verif}

This section asks whether naturalized text stays consistent with its source record. The record-based checker passes 87.2\% of English and 88.6\% of Korean outputs on the first generation, with a similar result for Spanish. After regeneration, a full-corpus re-audit finds no violations. The checker detects over 96.9\% of injected corruptions across all three languages, and manual inspection finds no semantic violations. After two regeneration rounds, about 8\% of items retain their canonical template rendering, so the corpus stays record-consistent without manual review of every item.

\vspace{-3.5mm}
\subsection{Language Portability and Record Reuse}
\label{sec:res-port}

This section asks whether the same records serve new languages and new question types without re-simulation. For languages, the Spanish realization module and verification lexicon were created in about 20 minutes and achieve a 94.3\% first-pass verification rate. Korean likewise changes only the language-specific components, and its probe performs within 7.1 points of the English means on the same meshes and facts. Reuse thus holds end to end, from simulation to a trained probe, without regenerating gold answers.

For question types, a fourth task on intervention effects is constructed directly from the stored records with the template mechanism of Sec.~\ref{sec:naturalization}, and a probe trained only on the original objectives performs near chance on it (27.5\% on a class-balanced subset, chance 33.3\%; \cite{extended3dtongueqa}). The records thus provide new supervision, not only new linguistic forms.

\vspace{-3.5mm}
\subsection{Utility and Geometry Grounding}
\label{sec:res-utility}

This section asks whether the supervision is learnable and whether what is learned depends on the paired geometry. Table~\ref{tab:results} reports all scores.

\noindent\textbf{Domain specificity and learnability.}
We first test whether the muscle-grounded \emph{why} and target-directed \emph{how} are already available from general-purpose language knowledge. Zero-shot GPT-5 Pro reaches only 7.2 Muscle EM, while the text-only model stays well below the geometry-conditioned probe on all three tasks. After training on our supervision, the 3D Muscle-Aware probe reaches 62.9 Muscle EM and outperforms the text-only and shuffled controls on Value and Direction. The supervision is therefore not recoverable from general knowledge alone but is learnable from paired geometry.

\noindent\textbf{Native 3D geometry and muscle-aware representation.}
We next test which input representation the probe needs. Replacing the 3D mesh with matched 2D views reduces Muscle EM from 62.9 to 42.1, so native 3D geometry carries information useful for recovering the generating muscle state. Replacing muscle-aware pretraining with geometry-oriented Mesh-AE pretraining lowers Muscle substantially while Direction changes only modestly and Value is slightly higher, so muscle-aware pretraining helps specifically in recovering the generating state rather than uniformly across tasks.

\noindent\textbf{Dependence on paired geometry.}
We then test whether the answers depend on the specific mesh. Shuffling the mesh while keeping the question and gold answer fixed drops Muscle EM from 62.9 to 2.2, lowers Value sharply, and brings Direction close to the text-only baseline. Muscle and Value predictions therefore rely on the paired geometry rather than textual patterns, and the Direction performance remaining after shuffling is largely explained by textual priors.

\noindent\textbf{Retrieval and extrapolation.}
Against nearest-neighbor answer transfer, the 1-NN baseline performs strongly on Value, but the unified probe clearly outperforms it on Muscle and Direction, so local geometric similarity suffices for many scalar geometry values whereas the generating state and target-directed correction require more than retrieval.

\noindent\textbf{One record set, two interfaces.}
We test whether the supervision is tied to language decoding. Task-specific readouts on the same frozen 3D representation exceed 87\% on all three tasks, and mesh shuffling sharply reduces their Muscle and Direction performance. The readouts receive explicit task information and, for Direction, the target activation vector, whereas the QA probe must interpret the question and generate an answer, so the same records support both flexible language-based QA and high-accuracy fixed-schema prediction.

\noindent\textbf{Open-ended quality.}
Finally, we test what training changes in free-form answers. On held-out question templates, GPT-5 Pro is more fluent on average, but the difference is not significant after correction, whereas the Muscle-Aware probe achieves substantially higher reference-based Factual Accuracy at 3.81 versus 2.23. Training on the generated supervision thus mainly improves accurate delivery of simulator-grounded information, even under unseen linguistic forms, rather than fluency.

\vspace{-3.5mm}
\section{Conclusion}
\vspace{-2mm}
\label{sec:conclusion}

We presented a simulator-based framework for constructing muscle-grounded supervision that goes beyond describing \emph{what} a tongue configuration looks like. By linking each geometry to its simulator-defined generating muscle state and target-directed adjustments, the framework provides supervision for the muscle-grounded \emph{why} and target-directed \emph{how}. Structured fact records preserve provenance from simulation to QA, while gold answers fixed before linguistic realization enable record-based verification. The same records can also support new languages and question types without re-simulation. In 3DTongueQA, this supervision is learnable from paired geometry and supports both natural-language QA and task-specific prediction.

The current instantiation remains a controlled testbed rather than a substitute for measured articulation. It uses one Badin anatomy with fixed topology and does not model jaw--lip coupling. Its stored activations are simulator-defined generating labels rather than uniquely identifiable physiological causes. This supervision is nonetheless fully traceable and extensible, so a richer biomechanical model can add finer generating factors, articulatory components, and intervention signals without changing the construction principle. 

Beyond the present tasks, the framework can serve as research infrastructure for learner-feedback models. Its records supply verifiable muscle-level targets and target-directed adjustments that such models must learn to express, its released pipeline lets researchers add feedback question types and languages without new simulation, and its 3D surfaces yield vocal-tract area functions that connect each muscle state to its acoustic consequence. The records can further be paired with rtMRI and EMA recordings for articulatory inversion. The construction principle is also not specific to the tongue, since any simulator whose inputs generate observable geometry can serve as a ground-truth generator. Future work will extend anatomical variation and articulatory coupling to support more detailed and actionable articulatory feedback. Code and templates are released under the MIT license and QA data under CC BY 4.0.


\vspace{-2mm}
\section*{Compliance with Ethical Standards}
\vspace{-2mm}
Three adult expert raters voluntarily assessed anonymized,
machine-generated outputs with informed consent, acting as evaluators
rather than subjects; no personal or identifying information was
collected, and no patients, minors, or physiological data were involved.

\vspace{-4mm}
\section*{Acknowledgment}
\vspace{-2mm}
This work was partly supported by Institute of Information \& communications Technology Planning \& Evaluation (IITP) grant funded by the Korea government (MSIT) (No. 2022-0-00621, RS-2022-II220621, Development of artificial intelligence technology that provides dialog-based multi-modal explainability) and (IITP-RS-2026-25547954, Artificial Intelligence Innovation Human Resources Development). This work was also supported by the Open Project Program of the State Key Laboratory of CAD\&CG (Grant No. A2612), Zhejiang University.

\vspace{-2mm}
\bibliographystyle{IEEEbib}
\bibliography{refs}

@article{hui2024qwen2,
  title={Qwen2.5 Technical Report},
  author={{Qwen Team} and Yang, An and Yang, Baosong and others},
  journal={arXiv preprint arXiv:2412.15115},
  year={2025}
}

@inproceedings{narayanan2011multimodal,
  title={A multimodal real-time {MRI} articulatory corpus for speech research},
  author={Narayanan, Shrikanth and Bresch, Erik and Ghosh, Prasanta Kumar and others},
  booktitle={Proc. Interspeech 2011},
  pages={837--840},
  year={2011}
}

@article{narayanan2014real,
  title={Real-time magnetic resonance imaging and electromagnetic articulography database for speech production research (TC)},
  author={Narayanan, Shrikanth and Toutios, Asterios and Ramanarayanan, Vikram and others},
  journal={J. Acoust. Soc. Am.},
  volume={136},
  number={3},
  pages={1307--1311},
  year={2014}
}

@inproceedings{azuma2022scanqa,
  title={{ScanQA}: {3D} question answering for spatial scene understanding},
  author={Azuma, Daichi and Miyanishi, Taiki and Kurita, Shuhei and Kawanabe, Motoaki},
  booktitle={Proc. CVPR},
  pages={19107--19117},
  year={2022}
}

@article{ma2022sqa3d,
  title={{SQA3D}: Situated question answering in {3D} scenes},
  author={Ma, Xiaojian and Yong, Silong and Zheng, Zilong and others},
  journal={arXiv preprint arXiv:2210.07474},
  year={2022}
}

@incollection{lloyd2012artisynth,
  title={{ArtiSynth}: A fast interactive biomechanical modeling toolkit combining multibody and finite element simulation},
  author={Lloyd, John E and Stavness, Ian and Fels, Sidney},
  booktitle={Soft tissue biomechanical modeling for computer assisted surgery},
  pages={355--394},
  year={2012}
}

@inproceedings{vogt2006efficient,
  title={Efficient {3D} finite element modeling of a muscle-activated tongue},
  author={Vogt, Florian and Lloyd, John E and Buchaillard, St{\'e}phanie and others},
  booktitle={Proc. Int. Symp. Biomedical Simulation},
  pages={19--28},
  year={2006},
  organization={Springer}
}

@inproceedings{badin1998three,
  title={A three-dimensional linear articulatory model based on {MRI} data},
  author={Badin, Pierre and Bailly, G{\'e}rard and Raybaudi, Monica and Segebarth, Christoph},
  booktitle={Proc. SSW 1998},
  pages={249--254},
  year={1998}
}

@inproceedings{gong2019spiralnet++,
  title={{SpiralNet++}: A fast and highly efficient mesh convolution operator},
  author={Gong, Shunwang and Chen, Lei and Bronstein, Michael and Zafeiriou, Stefanos},
  booktitle={Proc. ICCV Workshops},
  pages={4141--4148},
  year={2019}
}

@article{yang2025qwen3,
  title={Qwen3 technical report},
  author={Yang, An and Li, Anfeng and Yang, Baosong and others},
  journal={arXiv preprint arXiv:2505.09388},
  year={2025}
}

@misc{GPT-pro,
  author = {OpenAI},
title = {{ChatGPT} ({GPT-5 Pro}, {Aug. 12} version) [large language model]},
  year = {2026},
  howpublished = {\url{https://chat.openai.com/}},
  note = {Accessed Aug. 12, 2026}
}

@article{katz2015visual,
  title={Visual feedback of tongue movement for novel speech sound learning},
  author={Katz, William F and Mehta, Sonya},
  journal={Front. Hum. Neurosci.},
  volume={9},
  pages={612},
  year={2015}
}

@article{buchaillard2009biomechanical,
  title={A biomechanical model of cardinal vowel production: Muscle activations and the impact of gravity on tongue positioning},
  author={Buchaillard, St{\'e}phanie and Perrier, Pascal and Payan, Yohan},
  journal={J. Acoust. Soc. Am.},
  volume={126},
  number={4},
  pages={2033--2051},
  year={2009}
}

@article{takano2007mri,
  title={An {MRI} analysis of the extrinsic tongue muscles during vowel production},
  author={Takano, Sayoko and Honda, Kiyoshi},
  journal={Speech Commun.},
  volume={49},
  number={1},
  pages={49--58},
  year={2007}
}

@article{stavness2012biomechanical,
  title={Biomechanical modeling of {English} /r/ variants},
  author={Stavness, Ian and Gick, Bryan and Derrick, Donald and Fels, Sidney},
  journal={J. Acoust. Soc. Am.},
  volume={131},
  number={5},
  pages={EL355--EL360},
  year={2012}
}

@article{hillenbrand1995,
  title={Acoustic characteristics of {American English} vowels},
  author={Hillenbrand, James and Getty, Laura A and Clark, Michael J and Wheeler, Kimberlee},
  journal={J. Acoust. Soc. Am.},
  volume={97},
  number={5},
  pages={3099--3111},
  year={1995},
  doi={10.1121/1.411872}
}

@article{honda1996,
  title={Organization of tongue articulation for vowels},
  author={Honda, Kiyoshi},
  journal={Journal of Phonetics},
  volume={24}, number={1}, pages={39--52}, year={1996}
}

@article{compartmental2024,
  title={The Compartmental Tongue},
  author={Wrench, Alan A and others},
  journal={Journal of Speech, Language, and Hearing Research},
  year={2024},
  doi={10.1044/2024_JSLHR-23-00125}
}

@article{strycharczuk2025lingual,
  title={Dimensionality reduction in lingual articulation of vowels: Evidence from lax vowels in {Northern Anglo-English}},
  author={Strycharczuk, Patrycja and Kirkham, Sam and Gorman, Emily and Nagamine, Takayuki},
  journal={Language and Speech},
  volume={68},
  number={3},
  pages={689--721},
  year={2025},
  doi={10.1177/00238309251320581}
}

@incollection{flemming2009schwa,
  title={The phonetics of schwa vowels},
  author={Flemming, Edward},
  booktitle={Phonological Weakness in {English}: From {Old} to {Present-Day English}},
  editor={Minkova, Donka},
  pages={78--95},
  year={2009},
  doi={10.1007/978-0-230-29686-2_5}
}

@article{gomez2020fiberstrain,
  title={Analysis of fiber strain in the human tongue during speech},
  author={Gomez, Arnold D and Stone, Maureen L and Woo, Jonghye and Xing, Fangxu and Prince, Jerry L},
  journal={Computer Methods in Biomechanics and Biomedical Engineering},
  volume={23},
  number={8},
  pages={312--322},
  year={2020},
  doi={10.1080/10255842.2020.1722808}
}

@article{extended3dtongueqa,
      title={Simulate, record, verify: A language-portable framework for muscle-grounded articulatory {QA} (extended version)}, 
      author={Seungho Eum and Shantong Sun and Unsang Park},
      year={2026},
      journal={arXiv preprint arXiv:2608.23137},
      eprint={2608.23137},
      archivePrefix={arXiv},
      primaryClass={cs.CV},
      url={https://arxiv.org/abs/2608.23137}, 
}

@article{stavness2012automatic,
  title={Automatic prediction of tongue muscle activations using a finite element model},
  author={Stavness, Ian and Lloyd, John E and Fels, Sidney},
  journal={J. Biomech.},
  volume={45},
  number={16},
  pages={2841--2848},
  year={2012}
}

@inproceedings{tolpadi2018inverse,
  title={Inverse biomechanical modeling of the tongue via machine learning and synthetic training data},
  author={Tolpadi, Aniket A and Stone, Maureen L and Carass, Aaron and others},
  booktitle={Proc. SPIE Medical Imaging},
  year={2018}
}

@article{harandi2017variability,
  title={Variability in muscle activation of simple speech motions: A biomechanical modeling approach},
  author={Harandi, Negar M and Woo, Jonghye and Stone, Maureen and Abugharbieh, Rafeef and Fels, Sidney},
  journal={J. Acoust. Soc. Am.},
  volume={141},
  number={4},
  pages={2579--2590},
  year={2017}
}

@inproceedings{hong20233dllm,
  title={{3D-LLM}: Injecting the {3D} world into large language models},
  author={Hong, Yining and Zhen, Haoyu and Chen, Peihao and others},
  booktitle={Proc. NeurIPS},
  volume={36},
  year={2023}
}

@inproceedings{xu2024pointllm,
  title={{PointLLM}: Empowering large language models to understand point clouds},
  author={Xu, Runsen and Wang, Xiaolong and Wang, Tai and others},
  booktitle={Proc. ECCV},
  year={2024}
}

@inproceedings{johnson2017clevr,
  title={{CLEVR}: A diagnostic dataset for compositional language and elementary visual reasoning},
  author={Johnson, Justin and Hariharan, Bharath and van der Maaten, Laurens and others},
  booktitle={Proc. CVPR},
  pages={2901--2910},
  year={2017}
}

\twocolumn[{
  \begin{center}
    {\Large\bfseries Supplementary Material}\\[4pt]
    {\large Simulate, Record, Verify: A Language-Portable Framework
    for Muscle-Grounded Articulatory QA}
  \end{center}
  \vspace{10pt}
}]
\setcounter{section}{0}
\renewcommand{\thesection}{\Alph{section}}
% =====================================================================
% 3DTongueQA Supplementary Material — restructured concise draft
%
% Organization:
%   1) Dataset construction
%   2) Grounded QA construction
%   3) Experimental details and additional analyses
%
% Expected packages from the main manuscript:
%   amsmath, amssymb, booktabs, array, graphicx, multirow, url, tipa
%
% Author-only TODO items are placed after \endinput and do not appear in PDF.
% =====================================================================

\section{Instantiation Details for 3DTongueQA Construction}
\label{sec:supp_dataset}

The construction pipeline preserves provenance from simulator input to the
final linguistic instance as
\begin{equation}
\label{eq:supp_pipeline}
\mathbf{a}
\xrightarrow{\mathcal{S}}
\mathbf{M}
\xrightarrow{\mathcal{R}}
\mathbf{r}
\xrightarrow{\mathcal{G}}
(q,a)
\xrightarrow{\mathcal{N}}
(\widetilde q,\widetilde a),
\end{equation}
where $\mathbf{a}\in\mathbb{R}^{11}$ is a controlled muscle-activation
vector, $\mathcal{S}$ is the finite-element simulator, $\mathbf{M}$ is the
resulting fixed-topology tongue mesh, $\mathbf{r}$ is a structured fact
record, $\mathcal{G}$ deterministically constructs canonical QA, and
$\mathcal{N}$ changes only the linguistic surface form. The first four stages
determine the factual content, and a naturalized instance is retained only
after it passes a record-based faithfulness check, with the canonical
rendering kept when every attempt fails.

\subsection{Biomechanical Model, Mesh Generation, and Validity Filtering}
\label{sec:supp_generation}

We use the ArtiSynth \texttt{StableFemMuscleTongueDemo}, which implements the
Badin finite-element tongue model~\cite{lloyd2012artisynth,badin1998three}.
All samples share a fixed topology of $370$ surface vertices, $736$ triangular
faces, $948$ FEM nodes, and $740$ volumetric elements. Coordinates are
expressed in meters, where $Y{=}0$ is the midsagittal plane, $X$ runs from
anterior to posterior, and $Z$ points superiorly. The model is left--right symmetric.

The tongue is actuated by $11$ muscles in the fixed order GGP, GGM, GGA, STY,
GH, MH, HG, VERT, TRANS, IL, and SL. Each activation is bounded by $0.9$, and the
total activation is limited to $2.0$ for vowel-oriented configurations and
$2.3$ for consonant-oriented configurations. These values are simulator
controls rather than directly measured physiological activation levels.

Each activation vector is introduced through a minimum-jerk ramp
\begin{equation}
\label{eq:supp_minimum_jerk}
s(\tau)=10\tau^{3}-15\tau^{4}+6\tau^{5},
\end{equation}
executed over $1.0$~s with $24$ knots, followed by adaptive settling for
$0.2$--$1.5$~s. A watchdog records numerical solver status, element inversion,
per-element and global volume ratios, and residual and peak nodal velocities.
A run is labeled \texttt{FAILED\_NUMERICAL} when positions or velocities become
non-finite or the maximum nodal speed exceeds $10^{3}$~m/s. It is labeled
\texttt{INVALID\_PHYSICAL} when any FEM element is inverted, any per-element
volume ratio is non-positive, or the global volume ratio lies outside
$[0.60,1.40]$. A run is labeled \texttt{MARGINAL} when it fails the settling
criterion or triggers a softer velocity or volume warning. Only
\texttt{VALID} configurations are used for QA construction, training, and
evaluation.

\subsection{Reasoning-Objective-Driven Activation Sampling}
\label{sec:supp_sampling}

The activation pool combines eight subsets designed for distinct reasoning
relations rather than for corpus size alone. REST, SINGLE, PAIR, and TRIPLE
cover neutral, isolated, pairwise, and higher-order muscle states. ANCHOR,
NEIGHBOR, EFFORT, and SPACEFILL provide phoneme-associated postures, matched
single-muscle interventions, dose--response trajectories, and broader coverage
of the feasible activation space. Table~\ref{tab:supp_sampling_validity}
reports both the sampling rule and the number retained after biomechanical
screening.

\begin{table*}[t]
\centering
\caption{Activation sampling and biomechanical screening. $n_{\mathrm{act}}$
denotes the number of simultaneously active muscles. ``Retained'' includes
only configurations labeled \texttt{VALID}.}
\label{tab:supp_sampling_validity}
\setlength{\tabcolsep}{4.2pt}
\small
\resizebox{\textwidth}{!}{%
\begin{tabular}{lrrrp{10.2cm}}
\toprule
Subset & Screened & Retained & $n_{\mathrm{act}}$ & Sampling rule and supervision role \\
\midrule
REST
& 1 & 1 & 0
& All muscles inactive. It supplies the reference geometry and neutral or
abstention cases. \\
SINGLE
& 66 & 66 & 1
& Each of the $11$ muscles at
$\{0.15,0.30,0.45,0.60,0.75,0.90\}$, isolating individual-muscle effects. \\
PAIR
& 495 & 494 & 2
& All $55$ pairs on the $3{\times}3$ grid $\{0.3,0.6,0.9\}^{2}$. It supports
pairwise interaction analysis. \\
TRIPLE
& 3,000 & 2,998 & 3
& Three distinct muscles with amplitudes sampled from $[0.15,0.9]$ and
rescaled when the total budget is exceeded, extending coverage to higher-order
combinations. \\
ANCHOR
& 103,731 & 103,729 & 2--6
& Samples around $12$ vowel and $8$ consonant-class anchors. Approximately
$30\%$ interpolate between category centers to provide transition postures. \\
NEIGHBOR
& 58,060 & 58,059 & $\sim4$
& Changes exactly one muscle by $\pm0.15$ from an ANCHOR base, storing the
base index, changed muscle, and signed delta, forming matched counterfactual
pairs. \\
EFFORT
& 42,473 & 42,467 & 1--3
& Scales a fixed sparse muscle pattern over six global effort levels in
$[0.15,1.0]$, supplying explicit dose--response trajectories. \\
SPACEFILL
& 87,331 & 87,301 & 1--11
& Sparsity-biased sampling concentrated on two to five active muscles, with
nonzero amplitudes in $[0.15,0.9]$, broadening feasible-space coverage. \\
\midrule
\textbf{Total}
& \textbf{295,157} & \textbf{295,115} & --
& \textbf{The remaining $42$ configurations comprise $3$ marginal runs and
$39$ numerical failures, and no run is labeled invalid physical.} \\
\bottomrule
\end{tabular}%
}
\end{table*}

The stochastic budget targets proportions of $0.35$, $0.20$, $0.15$, and
$0.30$ for ANCHOR, NEIGHBOR, EFFORT, and SPACEFILL. Vowel-type vectors are
uniformly rescaled when the total budget is exceeded, whereas consonant-type
vectors reduce non-defining background muscles so that class-defining
constraints are preserved. Duplicate vectors are removed after rounding each
activation to four decimal places. Generation and global shuffling use fixed
seeds $0$ and $42$.

\noindent\textbf{Construction cost.}
The entire pipeline runs on a single desktop CPU, an Intel i5-13400F, with
no manual annotation at any stage. One configuration requires approximately
$43$~s of single-instance simulation with about $1$--$2$ cores and
$1$--$2$~GB of memory per headless instance, corresponding to roughly
$3{,}500$ CPU-hours for the full $295{,}157$-configuration pool, and running
$4$--$6$ instances in parallel reduces wall-clock time proportionally. Record
extraction, that is, feature computation from meshes, requires $10$--$20$
minutes per $10{,}000$ meshes. Once records are stored, deterministic
rendering of the full $891{,}156$-QA corpus takes $3$~min~$15$~s, so new
question types or languages never repeat the extraction step. Extending the corpus with a new
$10{,}000$-configuration distribution therefore costs approximately one day
on one desktop machine. Language naturalization with Qwen2.5-72B-Instruct on two NVIDIA H200 GPUs
processes approximately $10{,}000$ instances in $6.7$ minutes, and the full
corpus required $34.2$~h for English and $34.4$~h for Korean of wall-clock
time, including regeneration rounds. The record-based faithfulness check
passes $87.24\%$ of English and $88.55\%$ of Korean outputs on first
generation, that is, $2{,}789{,}567$ and $2{,}831{,}458$ of $3{,}197{,}604$,
rising cumulatively to $90.74\%$ and $92.32\%$ after one regeneration round
and $92.23\%$ and $93.73\%$ after a second. Only passing naturalized
instances are retained, and the remaining items keep their canonical
rendering.

\subsection{Phoneme Anchor Construction and Acoustic Plausibility}
\label{sec:supp_anchors}

The ANCHOR subset contains $12$ vowel postures and $8$ consonant classes.
Vowels are sampled as Gaussian perturbations around literature-informed
$11$-D targets, with noise applied only to nonzero target entries. MH is
additionally sampled from $\mathcal{U}(0.1,0.3)$ with probability $0.35$, and
vectors are uniformly rescaled to satisfy $\sum_m a_m\leq2.0$. Consonants are
specified by defining-muscle intervals and sparse background activation. Each
background muscle is activated with probability $0.35$ in
$\mathcal{U}(0,0.15)$, and MH uses $\mathcal{U}(0,0.25)$ with probability $0.5$.
Defining intervals overwrite the background, each activation is clipped to
$[0,0.9]$, and only non-defining muscles are reduced when the consonant budget
$\sum_m a_m\leq2.3$ is exceeded. The defining constraints are then rechecked
by rejection sampling. Tables~\ref{tab:supp_vowel_anchors} and
\ref{tab:supp_consonant_anchors} give the implemented settings.

\begin{table*}[t]
\centering
\caption{Vowel anchors. Only nonzero entries in the $11$-D target are shown, and
$\sigma$ denotes the Gaussian perturbation applied to those entries.}
\label{tab:supp_vowel_anchors}
\setlength{\tabcolsep}{4pt}
\footnotesize
\begin{tabular}{llcp{5.2cm}p{5.8cm}l}
\toprule
Category & IPA & $\sigma$ & Nonzero target activations & Primary articulatory rationale & Refs. \\
\midrule
High front & /i/ & 0.08
& GGP $0.50$, GGA $0.50$, GH $0.40$
& Forward and high tongue-body posture. & \cite{buchaillard2009biomechanical,honda1996} \\
Near-high front & /\textipa{I}/ & 0.12
& GGP $0.40$, GGA $0.38$, GH $0.30$
& Reduced-activation counterpart of /i/. & \cite{strycharczuk2025lingual} \\
Mid front & /e/ & 0.12
& GGP $0.42$, GGA $0.32$, GH $0.32$
& Intermediate front raising. & \cite{honda1996} \\
Open-mid front & /\textipa{E}/ & 0.12
& GGP $0.28$, GGA $0.42$, HG $0.18$
& Weaker fronting with mild tongue-body lowering. & \cite{strycharczuk2025lingual} \\
Low, /\ae/ merged & /a/ & 0.08
& GGA $0.45$, HG $0.55$
& Tongue-body depression with anterior support. & \cite{buchaillard2009biomechanical,takano2007mri} \\
Low back & /\textipa{A}/ & 0.10
& STY $0.28$, HG $0.55$
& Combined lowering and retraction. & \cite{honda1996,takano2007mri} \\
Open-mid back & /\textipa{O}/ & 0.12
& STY $0.38$, GH $0.20$, HG $0.42$
& Retracted posture with partial lowering. & \cite{takano2007mri} \\
Mid back & /o/ & 0.12
& GGP $0.38$, STY $0.42$, GH $0.38$
& Balanced back raising and anterior support. & \cite{takano2007mri} \\
High back & /u/ & 0.08
& GGP $0.42$, STY $0.50$, GH $0.40$, SL $0.30$
& Dorsum raising toward the velum. & \cite{buchaillard2009biomechanical,gomez2020fiberstrain} \\
Near-high back & /\textipa{U}/ & 0.12
& GGP $0.35$, STY $0.40$, GH $0.32$
& Reduced-activation counterpart of /u/. & \cite{strycharczuk2025lingual} \\
High central & /\textipa{1}/ & 0.12
& GGP $0.40$, GGA $0.25$, STY $0.20$
& Balanced fronting and retraction. & \cite{honda1996} \\
Schwa & /\textipa{@}/ & 0.14
& GGP $0.20$, GGA $0.18$
& Near-neutral posture with the widest within-class spread. & \cite{flemming2009schwa} \\
\bottomrule
\end{tabular}
\end{table*}

\begin{table*}[t]
\centering
\caption{Consonant-class anchors. Intervals specify defining constraints, and
upper bounds may act as exclusion constraints separating neighboring classes.}
\label{tab:supp_consonant_anchors}
\setlength{\tabcolsep}{4pt}
\footnotesize
\resizebox{\textwidth}{!}{%
\begin{tabular}{llp{5.8cm}p{6.3cm}l}
\toprule
Category & IPA & Defining activation intervals & Primary articulatory rationale & Refs. \\
\midrule
Velar stop & /k, g/
& STY $[0.70,0.90]$, GGP $[0.50,0.90]$, TRANS $[0.00,0.25]$
& Dorsum raising and retraction for velar closure. TRANS is capped to exclude
sibilant-like grooving. & \cite{harandi2017variability,gomez2020fiberstrain} \\
Velar nasal & /\textipa{N}/
& STY $[0.60,0.88]$, GGP $[0.45,0.85]$
& Slightly relaxed velar lingual posture. Nasality is extra-lingual. & interpolated from /k, g/ \\
Alveolar stop & /t, d/
& SL $[0.50,0.90]$, GGA $[0.30,0.70]$, VERT $[0.20,0.60]$,
TRANS $[0.00,0.18]$
& Tip/blade raising with TRANS capped to distinguish stops from /s/. & \cite{harandi2017variability} \\
Sibilant & /s/
& TRANS $[0.45,0.90]$, SL $[0.30,0.70]$, GGA $[0.30,0.70]$,
VERT $[0.20,0.55]$
& TRANS-induced midline grooving with the blade near the alveolar ridge. & \cite{gomez2020fiberstrain} \\
Lateral & /l/
& SL $[0.45,0.90]$, GGA $[0.25,0.60]$, IL $[0.00,0.30]$,
STY $[0.00,0.22]$
& Apical contact with limited backing. Lateral airflow itself is off-midsagittal. & \cite{compartmental2024} \\
Rhotic & /r/
& GGP $[0.40,0.80]$, GGA $[0.25,0.60]$, STY $[0.20,0.50]$
& GGP bunching with moderate retraction, separated from the velar range. & \cite{stavness2012biomechanical} \\
Postalveolar & /\textipa{S}, \textipa{Z}/
& TRANS $[0.35,0.75]$, GGA $[0.30,0.65]$, STY $[0.20,0.50]$,
SL $[0.20,0.55]$
& Grooving plus mild retraction places the constriction behind the alveolar ridge. & \cite{gomez2020fiberstrain} \\
Palatal glide & /j/
& GGP $[0.40,0.80]$, GGA $[0.40,0.80]$, GH $[0.20,0.50]$
& An /i/-like high-front posture sampled as a band. & \cite{buchaillard2009biomechanical} \\
\bottomrule
\end{tabular}%
}
\end{table*}

The /\ae/ target is merged into /a/ because the implemented model produces
near-identical midsagittal meshes for the two targets, with a vertex RMS of $0.9$~mm.
The lateral passage of /l/ and the surface groove of
/\textipa{S},\textipa{Z}/ are not directly represented by a midsagittal feature
set and should therefore be interpreted as literature-grounded anchor settings
rather than fully observed geometric properties.

\begin{figure}[t]
    \centering
    \includegraphics[width=0.46\linewidth]{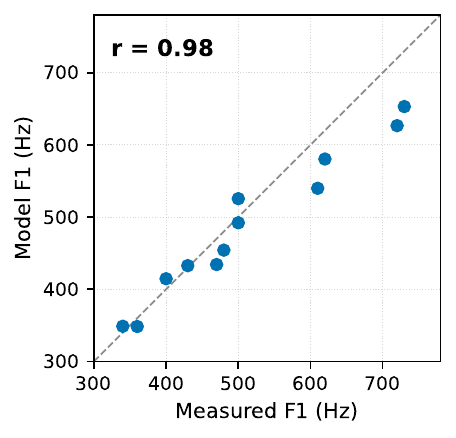}
    \hfill
    \includegraphics[width=0.46\linewidth]{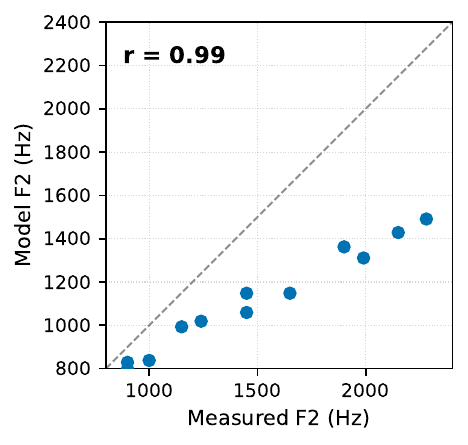}
    \caption{Measured versus model-derived vowel formants, $F_1$ with $r{=}0.98$
    and $F_2$ with $r{=}0.99$. Dashed lines denote identity. The relative vowel
    organization is preserved, although the absolute $F_2$ range is
    compressed.}
    \label{fig:supp_formant}
\end{figure}

For an acoustic plausibility check, a vocal-tract area function is estimated
from the simulated tongue, fixed palate, and Badin airway walls and evaluated
with a lossless concatenated-tube model. Category-level simulated formants are
compared with Hillenbrand measurements~\cite{hillenbrand1995}. Pearson
correlations are $r=0.98$ for $F_1$ and $r=0.99$ for $F_2$. The relative vowel
organization is retained, but the absolute $F_2$ range is compressed,
particularly for front vowels, because of the fixed lip boundary, single
anatomy, vocal-tract-length differences, and midsagittal area-function
approximation. This analysis supports relative articulatory--acoustic
plausibility rather than speaker-independent absolute calibration, as shown in
Fig.~\ref{fig:supp_formant}.

\subsection{Final Corpus Statistics, Splits, and Leakage Audit}
\label{sec:supp_corpus_stats}

Table~\ref{tab:supp_corpus_counts} separates the units used to describe the
released resource. The $295{,}157$ screened activation configurations include
$295{,}115$ \texttt{VALID} meshes retained for downstream construction. Each
retained item stores the $11$-D activation vector, $370{\times}3$ surface
vertices, $948{\times}3$ FEM nodes, shared topology, validity metadata, and any
applicable anchor, intervention, or trajectory links.

\begin{table}[t]
\centering
\caption{Configuration-level outcome of biomechanical screening.
Record- and job-level counts are reported separately in
Table~\ref{tab:supp_qa_counts}.}
\label{tab:supp_corpus_counts}
\setlength{\tabcolsep}{5pt}
\small
\begin{tabular}{lr}
\toprule
Unit & Count \\
\midrule
Activation configurations screened & 295,157 \\
Valid mesh configurations retained & 295,115 \\
Marginal configurations excluded & 3 \\
Numerical failures excluded & 39 \\
Invalid-physical configurations & 0 \\
\bottomrule
\end{tabular}
\end{table}
\noindent\textbf{Primary split and evaluation sampling.}
The primary sample-held-out split is constructed independently
within each of the eight sampling subsets at a
90/5/5 train, validation, and test ratio using seed 42,
yielding 265,611 training, 14,752 validation, and
14,752 test configurations. All QA instances and naturalized
variants derived from the same mesh remain in the same
partition. For the primary automatic evaluation, we sample
without replacement 20 test configurations from each of the
20 anchor categories, comprising 12 vowel and 8 consonant
classes. This yields 400 anchor-balanced test meshes.
Each mesh is paired with one active-muscle recovery,
one geometric-value prediction, and one target-directed
correction item, resulting in 1,200 automatically scored
items. Evaluation-set sampling also uses seed 42.
Split-integrity checks confirm that configuration identifiers,
rounded activation vectors, and exact mesh hashes do not
overlap across the train, validation, and test partitions.
The dataset-leakage-controlled anchor-held-out analysis uses a
separate evaluation set and is described in Sec.~\ref{sec:supp_anchor_holdout}.

\section{Grounded QA Construction}
\label{sec:supp_qa}

\subsection{Structured Fact Records and Deterministic QA Generation}
\label{sec:supp_facts}

Each fact record contains five groups of fields. The first holds the $11$
activations, active-muscle set, dominant muscle, and muscle count. The second
holds a $32$-D articulatory feature vector, rest-relative displacement,
constriction, dorsum height, doming, curvature, and landmark-derived
quantities. The third holds anterior, middle, and posterior motion
statistics, the fourth physical-validity metadata, and the fifth SINGLE,
PAIR, NEIGHBOR, EFFORT, phonetic-anchor, and target-correction links. The
$11$ activations, the $32$ articulatory features, the $9$ regional
displacement statistics, and the volume ratio together form the $53$ factual
fields referred to in the main text, and Table~\ref{tab:supp_fact_fields}
lists them by group. At
inference time, a trainable model receives a tongue mesh through a geometry
encoder, not the underlying activation vector. The stored activation is a
privileged simulator label defining the generating state of that sample, not
a uniquely identifiable physiological cause recoverable from geometry alone.
The active-muscle set used by fact construction, gold answers, and all
official scoring applies the threshold $a_m>10^{-4}$, and Direction gold
pairs require $|\Delta a_m|\geq0.05$. A coarser $n_{\mathrm{act}}$ summary
field with $a_m\geq0.05$ appears only in distribution metadata and is not
used in any gold answer or score.

\begin{table*}[p]
\centering
\caption{The $53$ factual fields of a fact record, namely the $11$ muscle
activations, the $32$-D articulatory feature vector, the $9$ regional
displacement statistics, and the volume ratio. $t$ and \texttt{xn} denote
normalized anterior--posterior positions with $0$ anterior and $1$
posterior, $d(t)$ is the palate-to-tongue distance function along the
palate, and heights \texttt{z} are normalized.}
\label{tab:supp_fact_fields}
\setlength{\tabcolsep}{4pt}
\footnotesize
\begin{tabular}{>{\raggedright\arraybackslash}p{3.0cm}>{\raggedright\arraybackslash}p{4.3cm}p{9.0cm}}
\toprule
Field & Full name & Definition \\
\midrule
\multicolumn{3}{@{}l}{\emph{Muscle activations, $11$ fields. Simulator-defined activation in $[0,0.9]$, the privileged generating state rather than an observable of the mesh.}} \\
\texttt{GGP} & genioglossus posterior & Raises and advances the tongue dorsum. \\
\texttt{GGM} & genioglossus medius & Raises the tongue body. \\
\texttt{GGA} & genioglossus anterior & Lowers the anterior tongue, forms a groove, and advances the tip. \\
\texttt{STY} & styloglossus & Retracts and elevates the dorsum. \\
\texttt{GH} & geniohyoid & Pulls the hyoid forward and upward. \\
\texttt{MH} & mylohyoid & Elevates the floor of the mouth and the tongue body. \\
\texttt{HG} & hyoglossus & Lowers and retracts the tongue body. \\
\texttt{VERT} & verticalis & Flattens the tongue by compressing it vertically. \\
\texttt{TRANS} & transversus & Narrows the tongue laterally, elongating it upward and forward. \\
\texttt{IL} & inferior longitudinal & Lowers the tip and shortens and retracts the tongue. \\
\texttt{SL} & superior longitudinal & Raises the tip and shortens the tongue. \\
\midrule
\multicolumn{3}{@{}l}{\emph{Distance function and constriction, $8$ fields}} \\
\texttt{cd\_min} & minimum constriction degree & Minimum of $d(t)$, the smallest gap between palate and tongue. \\
\texttt{cl\_t} & constriction location & Normalized position $t$ at which \texttt{cd\_min} occurs. \\
\texttt{gap\_mean} & mean palate--tongue gap & Mean of $d(t)$. \\
\texttt{gap\_std} & palate--tongue gap standard deviation & Standard deviation of $d(t)$. \\
\texttt{gap\_front} & front-region mean gap & Mean of $d(t)$ over $t<1/3$. \\
\texttt{gap\_mid} & mid-region mean gap & Mean of $d(t)$ over $1/3\leq t<2/3$. \\
\texttt{gap\_back} & back-region mean gap & Mean of $d(t)$ over $t\geq2/3$. \\
\texttt{constr\_width} & constriction width & Fraction of $t$ with $d(t)<1.5\times$\texttt{cd\_min}, the extent of the narrow constriction. \\
\midrule
\multicolumn{3}{@{}l}{\emph{Distance-function samples, $8$ fields}} \\
\texttt{df\_0} $\ldots$ \texttt{df\_7} & distance-function samples $d(t_i)$ & Palate-to-tongue distance sampled at eight positions from anterior, $i=0$, to posterior, $i=7$. \\
\midrule
\multicolumn{3}{@{}l}{\emph{Tongue landmarks, $10$ fields}} \\
\texttt{tip\_xn} & tongue-tip anteroposterior position & Normalized \texttt{xn} of the tongue tip. \\
\texttt{tip\_z} & tongue-tip height & Normalized height of the tongue tip. \\
\texttt{peak\_xn} & dorsal-peak anteroposterior position & Anteroposterior position of the highest dorsum point. \\
\texttt{peak\_z} & dorsal-peak height & Height of the highest dorsum point. \\
\texttt{h\_front} & front-region tongue height & Contour height at \texttt{xn}$=0.25$. \\
\texttt{h\_mid} & mid-region tongue height & Contour height at \texttt{xn}$=0.50$. \\
\texttt{h\_back} & back-region tongue height & Contour height at \texttt{xn}$=0.75$. \\
\texttt{mean\_z} & mean tongue height & Mean height of the midsagittal contour. \\
\texttt{com\_xn} & center-of-mass anteroposterior position & Mean \texttt{xn} of the contour points. \\
\texttt{com\_z} & center-of-mass height & Mean height of the contour points. \\
\midrule
\multicolumn{3}{@{}l}{\emph{Shape and curvature, $6$ fields}} \\
\texttt{arc\_len} & normalized arc length & Dorsal contour length divided by the palate arc length. \\
\texttt{tilt} & front-to-back tilt & First-order slope of the height profile from front to back. \\
\texttt{ant\_slope} & anterior slope & Slope of the contour segment anterior to the peak. \\
\texttt{post\_slope} & posterior slope & Slope of the contour segment posterior to the peak. \\
\texttt{curv\_peak} & peak curvature & Maximum curvature of the dorsal contour. \\
\texttt{doming} & doming & Convexity of the tongue above the chord from tip to root. \\
\midrule
\multicolumn{3}{@{}l}{\emph{Regional displacement relative to rest, $9$ fields, for the front, middle, and back regions}} \\
\texttt{front/mid/back\_dx} & regional anteroposterior displacement & Mean anteroposterior displacement of the region from the rest configuration. \\
\texttt{front/mid/back\_dz} & regional vertical displacement & Mean vertical displacement of the region from the rest configuration. \\
\texttt{front/mid/back\_mag} & regional displacement magnitude & Mean displacement magnitude of the region from the rest configuration. \\
\midrule
\multicolumn{3}{@{}l}{\emph{Physical validity, $1$ field}} \\
\texttt{vol\_ratio} & volume ratio & Volume of the deformed FEM mesh divided by the rest volume. \\
\bottomrule
\end{tabular}
\end{table*}

The main supervision is organized into Cause, State, and Change. Table
\ref{tab:supp_qa_taxonomy} summarizes the corresponding deterministic source
records and answer forms.

\begin{table*}[t]
\centering
\caption{Reasoning objectives and deterministic source records.}
\label{tab:supp_qa_taxonomy}
\setlength{\tabcolsep}{4.5pt}
\small
\resizebox{\textwidth}{!}{%
\begin{tabular}{llp{6.3cm}p{5.8cm}}
\toprule
Objective & Scenario family & Source record and reasoning target & Canonical answer form \\
\midrule
Cause
& Generating muscle state
& Stored activation vector and active-muscle set associated with the observed mesh
& Unordered set of canonical muscle identities. \\
Cause/State
& Isolated and interacting effects
& SINGLE, PAIR, and TRIPLE records describing individual and combined deformation patterns
& Muscle/effect attribution or structured interaction relation. \\
State
& Geometric and articulatory state
& Displacement, constriction, dorsum height, doming, curvature, landmarks, and normalized features
& Scalar value, geometric category, or constrained description. \\
State
& Physical consistency
& Volume ratio, element inversion, and settling metadata
& Validity or physical-relation label with supporting value. \\
Change
& Matched counterfactual
& NEIGHBOR base, changed muscle, signed activation delta, and regional geometric change
& Changed muscle, direction, and geometric consequence. \\
Change
& Dose--response
& EFFORT trajectory, effort level, and ordered geometric response
& Direction or monotonic relation. \\
Change
& Target-directed adjustment
& Current activation state, target-anchor state, and target phoneme
& Unordered set of muscle and direction pairs. \\
\bottomrule
\end{tabular}%
}
\end{table*}

\begin{figure}[t]
  \centering
  \begin{minipage}[b]{0.30\linewidth}\centering
    \includegraphics[height=2.5cm]{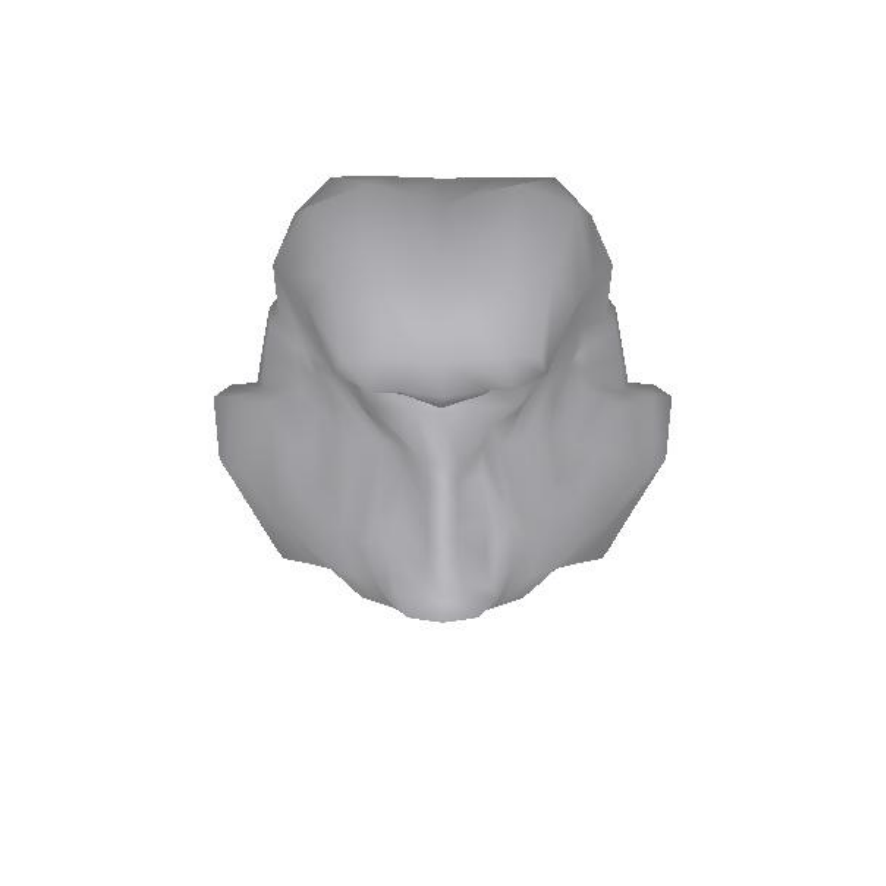}\\{\footnotesize anterior view}
  \end{minipage}\hfill
  \begin{minipage}[b]{0.30\linewidth}\centering
    \includegraphics[height=2.5cm]{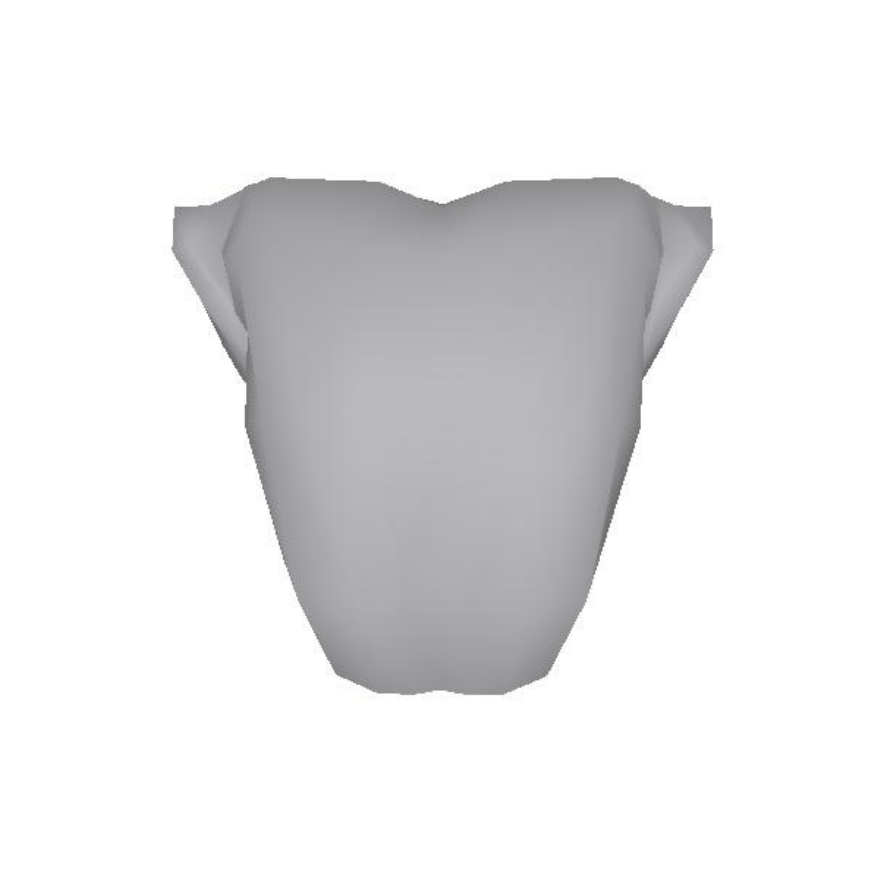}\\{\footnotesize dorsal view}
  \end{minipage}\hfill
  \begin{minipage}[b]{0.30\linewidth}\centering
    \includegraphics[height=2.5cm]{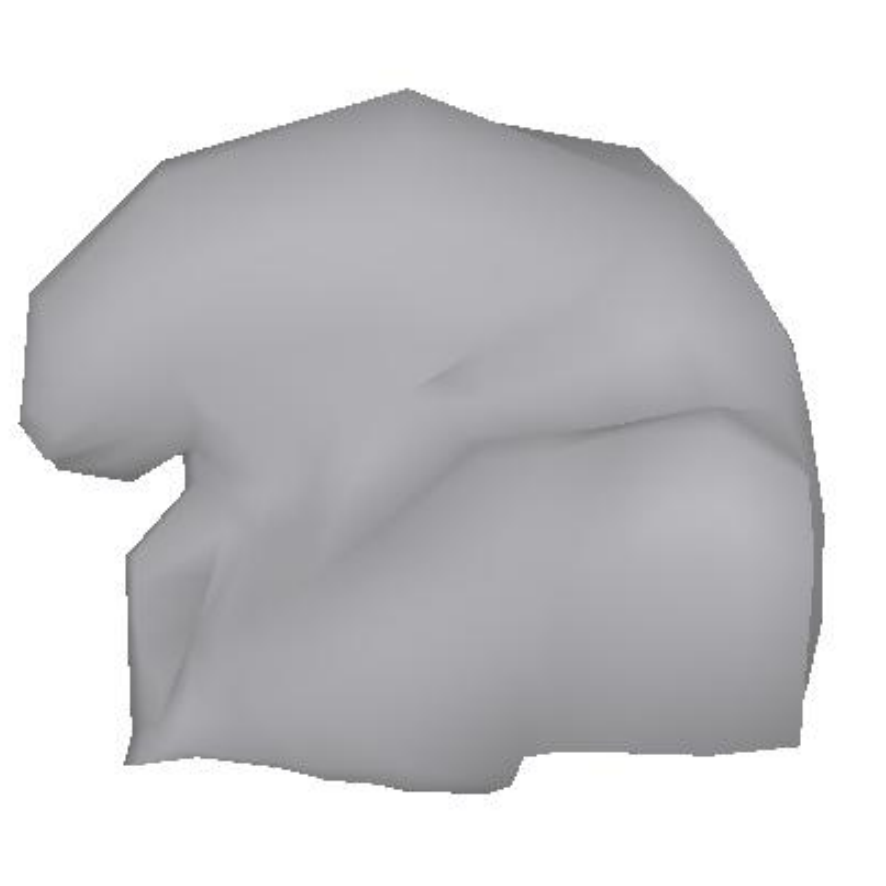}\\{\footnotesize lateral view}
  \end{minipage}\\[5pt]
  {\footnotesize\raggedright
   \textbf{Q:} How should the tongue be adjusted to approach the velar posture
   associated with /k/ or /g/?\par\smallskip
   \textbf{A:} Contract the genioglossus posterior (GGP) and styloglossus (STY)
   more, while relaxing the genioglossus anterior (GGA), geniohyoid (GH), and
   hyoglossus (HG). These changes move the activation pattern toward the
   simulator-defined velar anchor.\par}
  \caption{A target-directed instance constructed from a simulated tongue mesh
  and its current-state and target-anchor records. The same configuration is
  shown from three viewpoints. The mesh is rendered from the stored
  fixed-topology surface without any re-simulation.}
  \label{fig:qa_example}
\end{figure}

Twenty scenario definitions recombine these families into ordered multi-turn
conversations. Required fields are resolved from the fact record, and a
question is skipped or assigned an explicit abstention when its source fact is
undefined. Figure~\ref{fig:qa_example} shows a constructed target-directed
instance together with three views of its source mesh. Target-directed answers
encode only the simulator-defined muscle
increase or decrease directions toward an anchor, and they do not specify magnitude,
duration, a unique physiological control policy, or a clinically validated
treatment. Questions about whether an activation is uniquely identifiable from
geometry explicitly abstain because geometrically similar states may admit
alternative activation patterns.

\subsection{Naturalization, Faithfulness Verification, and QA Statistics}
\label{sec:supp_naturalization}

Track A contains canonical questions and answers rendered by deterministic
rules and serves as the factual reference. Track B contains naturalized variants
used for linguistic diversity. Naturalization uses Qwen2.5-72B-Instruct served
offline with vLLM in bfloat16 with tensor parallelism $2$ and prefix caching.
Three candidates are generated per conversation with temperature $0.7$,
top-$p=0.9$, and up to two retries, and after the final failed attempt the
canonical rendering is retained. English, Korean, and Spanish use
language-specific renderers and verification lexicons over the same
structured fact records. This demonstrates construction-level portability
rather than equal language difficulty.

\noindent\textbf{Third-language portability test with Spanish.}
Spanish was added after all simulation and training were complete by
translating the $20$ scenario templates and populating the verification
lexicon, and template translations were checked by back-translation.
Authoring took about $20$ minutes. A $400$-record sample was naturalized and
passed through the same record-based check, with $94.3\%$ passing on first
generation and a Wilson 95\% interval of $91.5$--$96.1\%$, whereas the
English and Korean rates are full-corpus figures. No Spanish probe was trained, so the Spanish
instantiation supports the portability claim at the construction level only.

\noindent\textbf{Manual check.}
One author manually compared $400$ naturalized QA pairs per language with
their source records, reading English and Korean directly and Spanish
through back-translation, and found no semantic violation in any language.

\noindent\textbf{Checker detection test.}
To measure the faithfulness checker independently of its own retained
output, we injected corruptions into $400$ passing naturalized outputs per
language, creating three corrupted variants of each output, $1{,}200$ per
language, namely a muscle identity replaced by another muscle, a
direction flipped, or the targeted concept, that is, the phoneme, feature,
or region, replaced. The record-based check detected $97.1\%$ of the
corrupted variants in English, $98.5\%$ in Korean, and $96.9\%$ in Spanish. Uncorrupted inputs are, by
construction of the sample, accepted at $100\%$.

\noindent\textbf{Corpus accounting.}
Counts are reported in naturalized conversation records. A
shape-description record contains three QA pairs, a physics-chain record
four, and a dose--response record one. The main run produces $891{,}156$
records per language, comprising $443{,}934$ shape description, $443{,}859$
physics chain, and $3{,}363$ dose--response records, and a further
$600$-record English pilot is excluded from all statistics. Faithfulness verification operates on
assistant-turn-level jobs, $3{,}197{,}604$ per language. The dose--response
family is deliberately small because its purpose is to pose non-monotonic
dose--response behavior as probe questions rather than to add training
mass. Training uses $240{,}000$ assistant turns per language, drawn from
variant-$0$ records of training meshes, which corresponds to approximately
$67{,}000$ conversation records or $7.5\%$ of the corpus, as listed in
Table~\ref{tab:supp_hyperparameters}. The remaining records are released
but were not required for the reported probes.

The faithfulness gate compares numerical values, canonical muscle names,
target phonemes, anatomical regions, directions, relation signs, and abstention
decisions against the source record. Accepted paraphrases have their annotation
spans recomputed. A full-corpus re-audit found no muscle-identity violation, no
omitted or altered gold numerical value, and no span-integrity error. Of $97$
turns containing an additional numeral, $92$ English cases were benign and the
remaining $5$ Korean artifacts were regenerated.

\begin{table}[t]
\centering
\caption{QA construction and naturalization statistics.}
\label{tab:supp_qa_counts}
\setlength{\tabcolsep}{5pt}
\small
\begin{tabular}{lr}
\toprule
Unit or audit quantity & Count \\
\midrule
Naturalized records per language & 891,156 \\
\quad Shape description & 443,934 \\
\quad Physics chain & 443,859 \\
\quad Dose--response & 3,363 \\
Records, both languages & 1,782,312 \\
Verification jobs per language & 3,197,604 \\
Muscle-identity violations & 0 \\
Gold numerical values omitted or altered & 0 \\
Turns with extraneous numerals & 97 \\
\quad Benign after inspection & 92 \\
\quad Regenerated artifacts & 5 \\
Span-integrity errors & 0 \\
\bottomrule
\end{tabular}
\end{table}

% [x] DONE: corpus accounting paragraph added; headline changed to 891,156
%     records per language (600-record EN pilot excluded).

\subsection{Post-Hoc Task Instantiation Without Regeneration}
\label{sec:supp_posthoc_task}

To demonstrate that the stored records support new deterministic tasks
without regenerating the corpus, we define an \emph{intervention-effect}
task after all simulation and training were complete. Given a NEIGHBOR
record linking a base configuration, a single changed muscle, and its
signed $\pm0.15$ delta, the question asks whether a named geometric
feature increases, decreases, or remains unchanged, and the gold answer
is computed by comparing the stored feature values of the two meshes
under the same dead bands as geometric-value scoring. Only pairs whose
base and neighbor configurations both lie in the test partition are used.
Generating all $990$ admissible instances and sampling $400$ of them,
balanced over muscles and features with a class distribution of $360/20/20$
for no-change, increase, and decrease, required $3.3$~seconds of wall-clock
time and zero additional simulation.

As a secondary analysis, the trained 3D Muscle-Aware unified-QA model is
evaluated on this task zero-shot, with free-form answers mapped
semantically to the three classes. It reaches $59.8\%$ on the
$400$-instance set, below the $90.0\%$ majority-class baseline, and
$27.5\%$ on a class-balanced $120$-instance subset, near the $33.3\%$
chance level. The new task is therefore not solved by the existing
supervision, indicating that post-hoc instantiated tasks can supply
complementary supervision rather than repackaging the original three
objectives.

\section{Experimental Details and Additional Analyses}
\label{sec:supp_experiments}

\subsection{Models, Training Hyperparameters, and Evaluation Protocol}
\label{sec:supp_training}

\noindent\textbf{Primary architecture.}
The primary 3D encoder uses SpiralNet++~\cite{gong2019spiralnet++}, vendored
at commit \texttt{96df4d8}, on rest-relative displacement of the fixed-topology
$370$-vertex surface. Two spiral-convolution blocks with widths $[16,16]$ and
QEM pooling factors $[4,4]$ produce the hierarchy
$370\rightarrow93\rightarrow24$. The final level supplies $24$
local tokens of dimension $16$, while a linear map produces a $32$-D global
latent. During muscle-aware pretraining, a sigmoid regression head predicts the
$11$-D activation vector with mean L1 loss, and the head is discarded before
QA training.

\begin{figure*}[t] \centering \includegraphics[width=\textwidth]{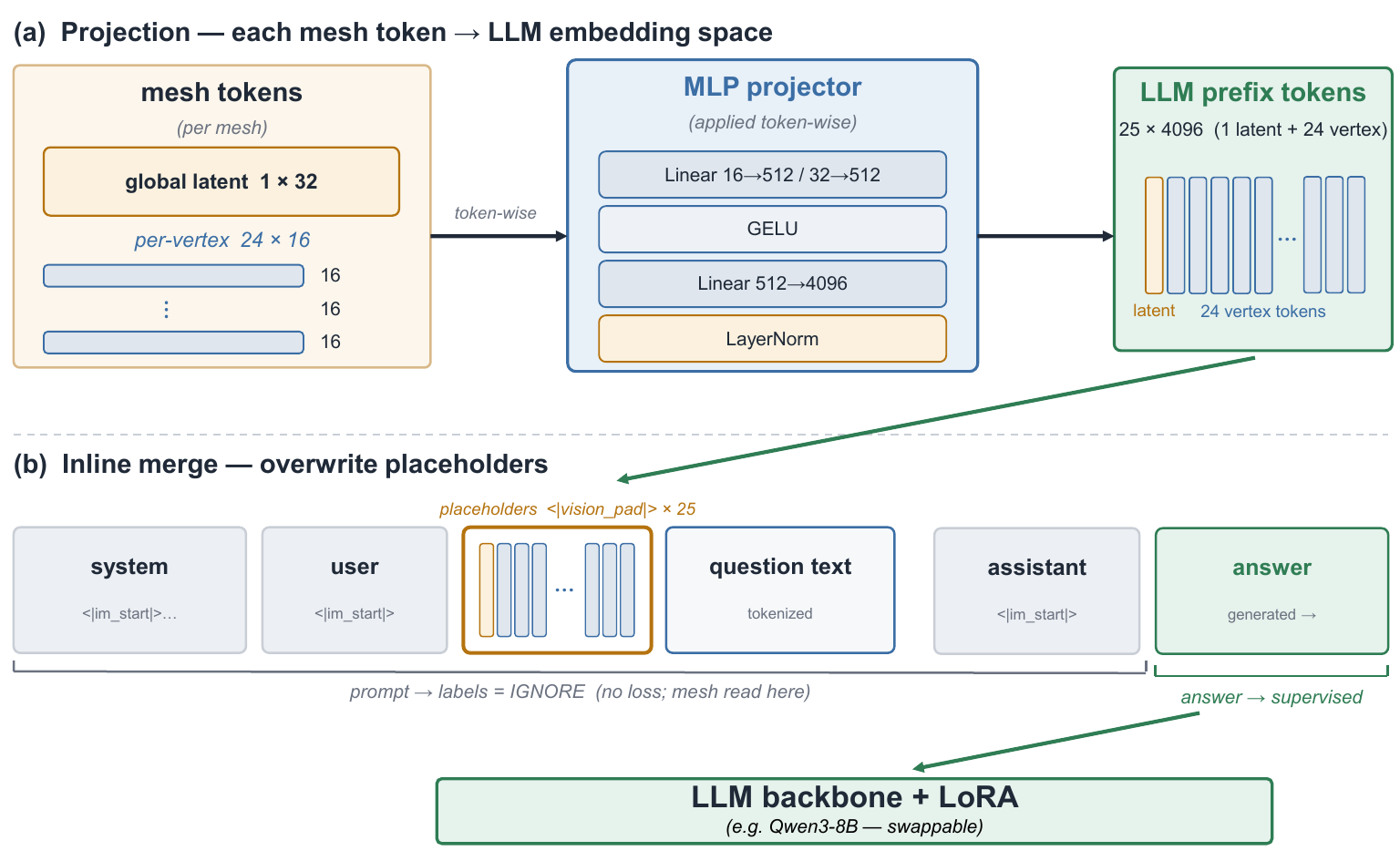} \caption{Overview of the projection pipeline connecting the mesh encoder to the language model. The encoded 3D tongue-mesh representation is converted into a compact set of projected tokens and provided to the LLM as a visual prefix for grounded articulatory question answering.} \label{fig:projection_overview} \end{figure*}

The frozen encoder outputs are z-score standardized and projected into the
$4096$-D Qwen3-8B embedding space through separate global and local MLPs of
widths $32\rightarrow512\rightarrow4096$ and
$16\rightarrow512\rightarrow4096$. One global and $24$ local
vectors form a $25$-token mesh prefix placed immediately before the question.
Qwen3-8B and the mesh encoder remain frozen, only the projector and LoRA
adapters are updated, and loss is applied only to assistant-answer tokens. Figure~\ref{fig:projection_overview} illustrates the overall projection pipeline from the mesh encoder to the LLM. 

\begin{table*}[t]
\centering
\caption{Training settings for encoder pretraining and downstream QA.}
\label{tab:supp_hyperparameters}
\setlength{\tabcolsep}{5pt}
\small
\begin{tabular}{p{2.6cm}p{3.6cm}p{9.1cm}}
\toprule
Stage & Setting & Value \\
\midrule
Encoder pretraining
& Input / target
& Rest-relative displacement of size $370{\times}3$ in meters. $11$-D activation in $[0,1]$. \\
Encoder pretraining
& Loss / optimizer
& Mean L1. Adam, learning rate $10^{-3}$, weight decay $0$,
$(\beta_1,\beta_2)=(0.9,0.999)$. \\
Encoder pretraining
& Schedule / batch
& StepLR with $\gamma=0.99$ per epoch. Batch $256$ for training and $64$ for validation. FP32. \\
Encoder pretraining
& Split / checkpoint
& Subset-stratified $265{,}611/14{,}752/14{,}752$ train/validation/test split,
seed $42$, with the checkpoint selected by validation L1. \\
\midrule
QA training
& Backbone / trainable modules
& \texttt{Qwen/Qwen3-8B}, bfloat16, thinking disabled. Projector in FP32 and LoRA trainable. Encoder and LLM frozen. \\
QA training
& LoRA
& Rank $16$, $\alpha=32$, dropout $0.05$. Applied to $q,k,v,o$ projections and MLP projections. \\
QA training
& Optimizer / learning rates
& AdamW. Projector $5{\times}10^{-4}$ with weight decay $10^{-2}$. LoRA $2{\times}10^{-4}$ with weight decay $0$. \\
QA training
& Schedule / batch / length
& Cosine schedule with $5\%$ warmup. Effective batch size $16$. $10$ epochs. Maximum length $2{,}048$. Gradient clipping at $1.0$. \\
QA training
& Data / selection
& $240{,}000$ naturalized assistant turns, approximately $67{,}000$ conversation records, sampled from training meshes, variant $0$, seed $42$. Checkpoint selected by validation informative-number F1. \\
QA training
& Fact weighting
& Half of stored fact spans are sampled per instance. Target fact-loss share $\tau=0.6$, weight clipped at $20$, and previous-turn fact spans masked in the input with probability $0.3$. \\
\bottomrule
\end{tabular}
\end{table*}

\noindent\textbf{Training cost.}
Training one unified-QA seed, that is, projector and LoRA fine-tuning of
Qwen3-8B on the frozen encoder, requires approximately three days of
wall-clock time on two NVIDIA H200 GPUs, so the reported three-seed results
correspond to roughly nine H200-pair-days per model variant.

\noindent\textbf{Matched controls.}
Table~\ref{tab:supp_controls} summarizes the controls. Unless noted, each
trainable LLM variant uses the same QA data budget, projector/LoRA training
recipe, and split, and only the frozen input front-end differs.

\begin{table*}[t]
\centering
\caption{Matched controls used for grounded validation.}
\label{tab:supp_controls}
\setlength{\tabcolsep}{4.5pt}
\small
\begin{tabular}{p{3.0cm}p{4.7cm}p{7.2cm}}
\toprule
Model & Input/front-end & Purpose and matched difference \\
\midrule
Text-only
& No mesh encoder, projector, or prefix tokens
& Tests what can be learned from QA language and answer priors alone. \\
2D Muscle-Aware
& Parameter-matched shared CNN over front, left, and superior renders
& Pretrained on the same $11$-D activation target. It tests rendered 2D views against native 3D geometry. \\
3D Mesh-AE
& Same SpiralNet++ encoder hierarchy, pretrained by mesh-displacement reconstruction
& Tests generic geometry-oriented pretraining against muscle-aware pretraining. \\
Task-specific structured readouts
& Same frozen 3D encoder and projected global/local mesh features, with Qwen3-8B replaced by an $11$-label active-set head, a feature-conditioned scalar head, or a per-muscle three-class direction head
& Tests efficient schema-specific recovery of simulator-defined muscle state and geometric state without language decoding. \\
GPT-5 Pro
& Zero-shot external reference given the raw 3D mesh file together with the question text
& Tests whether a general-purpose commercial model can solve the task without domain-specific training. \\
\bottomrule
\end{tabular}
\end{table*}

\begin{table}[t]
\centering
\caption{Effect of mesh shuffling while keeping the question and gold
answer fixed. Mean $\pm$ standard deviation over three seeds is reported
where available.}
\label{tab:shuffle}
\setlength{\tabcolsep}{4pt}
\small
\resizebox{\columnwidth}{!}{%
\begin{tabular}{lccc}
\toprule
Model & Muscle EM & Geom. Value Acc. & Direction EM \\
\midrule
3D Muscle-Aware + LLM & 2.2 $\pm$ 0.5 & 31.2 $\pm$ 0.2 & 40.9 $\pm$ 0.7 \\
3D Mesh-AE + LLM      & 2.8 $\pm$ 0.4 & 34.9 $\pm$ 0.3 & 39.7 $\pm$ 0.5 \\
2D Muscle-Aware + LLM & 2.2 $\pm$ 0.3 & 30.0 $\pm$ 0.7 & 38.5 $\pm$ 0.8 \\
Task-specific structured readouts & 1.2 $\pm$ 0.0 & 41.6 $\pm$ 0.6 & 1.8 $\pm$ 0.0 \\
3D Muscle-Aware + LLM, Korean & 2.8 & 30.8 & 36.8 \\
\bottomrule
\end{tabular}%
}
\end{table}

\noindent\textbf{Task-specific structured readouts.}
To test whether the supervision in 3DTongueQA is tied to language decoding, we retain
the frozen 3D Muscle-Aware encoder, stored feature normalization, and projected
global/local mesh features, but remove Qwen3-8B and its LoRA adapters. A
lightweight token reducer operates on the resulting one global and $24$ local
tokens while preserving their fixed topology-consistent order. The active-set
readout outputs $11$ muscle logits and is optimized with binary cross entropy
using the simulator-defined labels $a_m>10^{-4}$, and a single probability
threshold is selected on validation data and frozen for test. A separate geometric
readout receives the reduced mesh representation together with an explicit
identifier for one of the six evaluated geometric features and predicts one
scalar. The feature identifier is supplied directly rather than inferred from
a natural-language question. The direction readout uses the same frozen encoder with a newly initialized
MLP bridge, without copying LLM projector weights, and receives the
flattened mesh representation concatenated with the target anchor's
$11$-dimensional prototype activation vector from the anchor definitions. No
discrete anchor identifier and no gold activations are given as input. It
predicts, for each of the $11$ muscles, a three-way movement class of
no-change, increase, or decrease with per-muscle cross entropy. Gold classes are derived from the
nearest-mesh-corrected adjustment toward the target anchor under the
$|\Delta a_m|\geq0.05$ dead band. The predicted
set of muscle and direction pairs is formed from the non-neutral classes and scored with the same Direction EM as the unified model. The readouts are evaluated on the identical
400 Muscle, 400 Value, and 400 Direction items used by the unified-QA
model. Checkpoint selection uses validation data only.

\noindent\textbf{1-NN answer-transfer baseline.}
The retrieval baseline uses only the training partition as its bank. Each
test mesh is represented by its raw rest-relative vertex-displacement vector
of size $370{\times}3$ with fixed vertex correspondence and no learned
features, its nearest training mesh is found by Euclidean distance, and the
stored answer of that mesh for the matching question type, queried feature,
and target anchor where applicable is copied verbatim. No decoder is fitted. Scored on the
same $1{,}200$ items, it reaches $44.0$ Muscle EM, $97.2$ Geometric Value
Accuracy, and $47.3$ Direction EM.

\begin{table*}[t]
\centering
\caption{Six scalar mesh-derived properties used for geometric-value
evaluation, with the corresponding fields of
Table~\ref{tab:supp_fact_fields}. Each question requests one property.}
\label{tab:supp_geometric_features}
\setlength{\tabcolsep}{4.5pt}
\small
\begin{tabular}{p{3.8cm}lp{9.6cm}}
\toprule
Feature & Field & Interpretation \\
\midrule
Maximum-constriction location & \texttt{cl\_t}
& Normalized anterior--posterior position along the palate at which the tongue--palate distance is smallest. $0$ denotes anterior and $1$ posterior. \\
Minimum constriction degree & \texttt{cd\_min}
& Minimum tongue--palate distance at the constriction point. Smaller values indicate closer approximation or contact. \\
Dorsum-peak location & \texttt{peak\_xn}
& Normalized anterior--posterior position of the highest point on the tongue dorsum, which may differ from the maximum-constriction location. \\
Dorsum-peak height & \texttt{peak\_z}
& Vertical height of the highest dorsum point relative to the anterior-palate reference. \\
Doming & \texttt{doming}
& Elevation of the dorsum above the line joining the anterior and posterior tongue references. Larger values indicate a more arched profile. \\
Anterior--posterior dorsum slope & \texttt{tilt}
& Fitted anterior--posterior slope of the dorsum. A positive value indicates that the posterior dorsum is higher than the anterior dorsum. \\
\bottomrule
\end{tabular}
\end{table*}

\noindent\textbf{Evaluation protocol.}
The primary automatic evaluation uses 400 anchor-balanced
sample-held-out meshes and 1,200 items covering active-muscle
recovery, geometric-value prediction, and target-directed
correction. Forced-prefix cloze generation uses a target-free lead-in and at
most $128$ continuation tokens. Muscle EM requires an exact active-set match.
For geometric-value prediction, each item requests one of the six properties in
Table~\ref{tab:supp_geometric_features}. The 400 Value items are distributed as
evenly as possible across feature types, with 66 or 67 items per feature. For
feature $f$, we compute
\begin{equation}
A_f=\frac{1}{N_f}\sum_{i:f_i=f}
\mathbf{1}\bigl(|\hat v_i-v_i|\leq\delta_f\bigr),
\end{equation}
where $\delta_f$ is $1.0$~mm for metric-valued quantities, $0.1$ for
normalized quantities, and $0.03$ for the minimum constriction degree
\texttt{cd\_min}, and
region-movement facts use a $0.5$~mm displacement threshold. Geometric Value Accuracy is the
unweighted macro-average
$\frac{1}{6}\sum_{f=1}^{6}A_f$. The unified-QA score uses the scalar parsed from
the generated continuation, while the structured readout outputs the scalar
directly. Direction EM requires an exact match to the unordered set of muscle and
direction pairs. The automatic Direction parser
maps free-form answers to this set by accepting arrows, ``contract'' or
``increase'' as $\uparrow$, ``relax'' or ``decrease'' as $\downarrow$,
canonical muscle names and abbreviations, and Korean SOV phrasings. English
2D, 3D Mesh-AE, and 3D Muscle-Aware unified-QA results and the structured
readouts are averaged over three seeds. The structured readouts use the
identical 400 test items and scoring definitions. Human evaluation uses the same $30$ English prompts across models,
while the Korean model uses a separate Korean set.

\subsection{Additional Automatic and Human Analyses}
\label{sec:supp_additional}

\noindent\textbf{Decoder-agnostic utility.}
The task-specific active-set readout reaches $88.7\pm0.7$ Muscle EM, compared
with $62.9\pm9.2$ for the unified QA model on the same active-set recovery
task. The feature-conditioned scalar readout reaches $87.5\pm1.1$ macro
Geometric Value Accuracy, compared with $74.0\pm0.2$ for unified QA, and the
Direction readout reaches $93.3\pm1.0$ Direction EM, compared with
$65.9\pm4.7$. Per-muscle F1 of the active-set readout is at least $0.995$
for all $11$ muscles across the three seeds, with a macro F1 of $0.998$, and
GGP, GGM, GGA, STY, VERT, and IL reach $1.0$ in every seed. The geometric structured
readout receives the requested feature as an explicit identifier and predicts
one scalar, whereas the unified model must infer the requested quantity from
the natural-language question and generate the answer textually. These results
therefore show that the deterministic supervision supports strong specialized
readouts in addition to heterogeneous natural-language QA, and they are not
treated as a like-for-like comparison of language understanding.

\noindent\textbf{Direct-regression baseline.}
To test whether the active set is trivially recoverable from the pretrained
representation, we threshold the pretraining activation-regression head
directly, with no QA or readout training, from a single pretraining
checkpoint. Under the official $a>10^{-4}$ active-set definition this yields
$9.8$ Muscle EM with $70.5$ macro F1 and an activation MAE of $0.026$, and
$30.8$ under a coarser $0.02$ threshold, both far below the trained
active-set readout at $88.7\pm0.7$. The active set is therefore not a
trivial re-reading of the pretraining target, and readout training extracts
substantial additional information from the frozen representation.

\noindent\textbf{Set-similarity metrics for unified Muscle predictions.}
Because exact set match is a strict criterion, we additionally rescore the
stored unified-model Muscle predictions with per-muscle macro-F1 and
macro-Jaccard. On the anchor-balanced $400$-item set, the three seeds obtain
macro-F1 $86.4/83.3/89.6$ and macro-Jaccard $79.3/74.4/83.0$ against EM
$66.5/52.5/69.8$, indicating that most errors are near-misses that differ in
one or two muscles rather than wholesale failures. Under seed $42$, the
weakest muscles are IL, GGM, and TRANS with F1 of $0.63$, $0.70$, and
$0.85$, while GGP, GGA, STY, and SL exceed $0.97$.

\begin{table}[t]
\centering
\caption{Complementary decoder roles on the identical
400-mesh anchor-balanced evaluation set. Both settings use
the frozen 3D Muscle-Aware representation. The geometric
structured readout receives an explicit feature identifier, and
its score is the unweighted macro-average over six feature
types. Mean $\pm$ standard deviation over three seeds is
shown where available.}
\label{tab:structured_decoder}
\setlength{\tabcolsep}{4.5pt}
\small
\resizebox{\columnwidth}{!}{%
\begin{tabular}{lccc}
\toprule
Readout & Muscle EM & Geom. Value Acc. & Direction EM \\
\midrule
Unified Qwen3-8B QA
& 62.9\,$\pm$\,9.2
& 74.0\,$\pm$\,0.2
& 65.9\,$\pm$\,4.7 \\
Task-specific structured readouts
& \textbf{88.7\,$\pm$\,0.7}
& \textbf{87.5\,$\pm$\,1.1}
& \textbf{93.3\,$\pm$\,1.0} \\
\bottomrule
\end{tabular}%
}
\end{table}

\noindent\textbf{Dataset-leakage-controlled anchor-held-out stress test.}
\label{sec:supp_anchor_holdout}
This secondary analysis asks whether the learned mapping transfers to anchor
definitions removed from training. A naive single-anchor holdout is not
meaningful because nearby anchors can act as near duplicates. Removing /o/
alone leaves a 1-nearest-neighbor Muscle EM of $99.3$, and approximately $80\%$
of its nearest neighbors are retained ANCHOR samples.

For each anchor category $c$, let $\bar{\mathbf a}_c\in\mathbb{R}^{11}$ denote
its mean activation vector and define
\begin{equation}
 d(c,c')=\lVert\bar{\mathbf a}_c-\bar{\mathbf a}_{c'}\rVert_2.
\end{equation}
Candidate sets must satisfy three performance-independent rules. First,
\emph{near-anchor closure} requires every held-out anchor to be at least
$d=0.2$ from the retained inventory, and closer twins must be removed jointly.
Second, candidates whose labels or aliases appear in prescriptive training text
are excluded. This removes \texttt{i\_front}, \texttt{a\_low},
\texttt{u\_back}, \texttt{alveolar\_td}, \texttt{sibilant\_s},
\texttt{velar\_kg}, and, after alias expansion, \texttt{palatal\_j}.
Third, among the admissible candidates, a maximin rule selects the categories
with the largest minimum distance to the retained inventory. No final-model
score is used in selection.

For vowels, \{\texttt{e\_mid},\texttt{ih\_lax}\} is inadmissible because
\texttt{ih\_lax} is within $d=0.184$ of the retained \texttt{i\_front}, so the
only admissible pair is \{\texttt{o\_mid},\texttt{uh\_lax}\}, corresponding
to /o/ and /\textipa{U}/. For consonants, \texttt{velar\_ng} lies within
$d=0.126$ of \texttt{velar\_kg}, while the remaining admissible minimum
distances are $0.516$ for \texttt{postalv\_sh}, $0.414$ for
\texttt{lateral\_l}, and $0.308$ for \texttt{rhotic\_r}. The rule therefore
selects /\textipa{S},\textipa{Z}/ and /l/.

\begin{table}[t]
\centering
\caption{Performance-independent selection of the four held-out anchor
categories. Distances are computed between mean $11$-D activation vectors.}
\label{tab:supp_anchor_selection}
\setlength{\tabcolsep}{3.5pt}
\small
\resizebox{\columnwidth}{!}{%
\begin{tabular}{llcl}
\toprule
Type & Candidate & Minimum distance & Decision \\
\midrule
Vowel pair & \texttt{e\_mid}+\texttt{ih\_lax} & 0.184 & Rejected, near \texttt{i\_front} \\
Vowel pair & \texttt{o\_mid}+\texttt{uh\_lax} & $\geq0.2$ & Selected \\
\midrule
Consonant & \texttt{palatal\_j} & 0.168 & Rejected, text leak and near twin \\
Consonant & \texttt{velar\_ng} & 0.126 & Rejected, near \texttt{velar\_kg} \\
Consonant & \texttt{postalv\_sh} & 0.516 & Selected, rank 1 \\
Consonant & \texttt{lateral\_l} & 0.414 & Selected, rank 2 \\
Consonant & \texttt{rhotic\_r} & 0.308 & Admissible, not selected \\
\bottomrule
\end{tabular}%
}
\end{table}

The final held-out set is /o/, /\textipa{U}/, /l/, and
/\textipa{S},\textipa{Z}/. We remove their direct ANCHOR configurations,
interpolations involving a held-out endpoint, derived NEIGHBOR configurations,
all QA from those source meshes, and prescriptive QA using the held-out target
anchors from both encoder pretraining and QA training. Evaluation uses $500$
direct-anchor meshes balanced at $125$ per category. The exact meshes are
absent from both training conditions, but the full-inventory model may
observe other samples generated from the same definitions, whereas the
anchor-held-out model may not. This protocol controls dataset-side leakage
only, since phonetic world knowledge already present in the pretrained base LLM is not, and
cannot be, removed by data exclusion, which is why we refer to the protocol
as dataset-leakage-controlled.

\begin{table}[t]
\centering
\caption{Anchor-held-out stress test on an identical balanced $500$-mesh set.}
\label{tab:supp_anchor_results}
\setlength{\tabcolsep}{3.5pt}
\scriptsize
\resizebox{\columnwidth}{!}{%
\begin{tabular}{lccc}
\toprule
Training condition & Muscle EM & Geom. Value Acc. & Direction EM \\
\midrule
Full anchor inventory & 61.8 & 47.8 & 50.1 \\
Four anchors held out & 49.7 & 44.8 & 49.4 \\
\midrule
Retained performance & 80.4\% & 93.7\% & 98.6\% \\
\midrule
1-NN, held-out bank & 30.0 & 62.2 & 28.0 \\
\bottomrule
\end{tabular}%
}
\end{table}

The larger reduction in Muscle EM indicates greater sensitivity of exact
activation-set recovery to anchor-specific supervision, while Geometric Value Accuracy
and Direction EM are less affected. The 1-NN answer-transfer baseline, with
its bank restricted to the same held-out training partition, reaches
$30.0$/$62.2$/$28.0$ on the same $500$ meshes. Retrieval remains strong on
Geometric Value Accuracy, whose answers are deterministic functions of mesh
geometry, but collapses on Muscle EM and Direction EM, where the held-out
probe retains $49.7$ and $49.4$. The result demonstrates substantial but
incomplete transfer within the same simulator, anatomy, and topology, and it
is not interpreted as sim-to-real or unseen-anatomy generalization.

\begin{figure}[H]
\centering
\includegraphics[width=0.58\columnwidth]{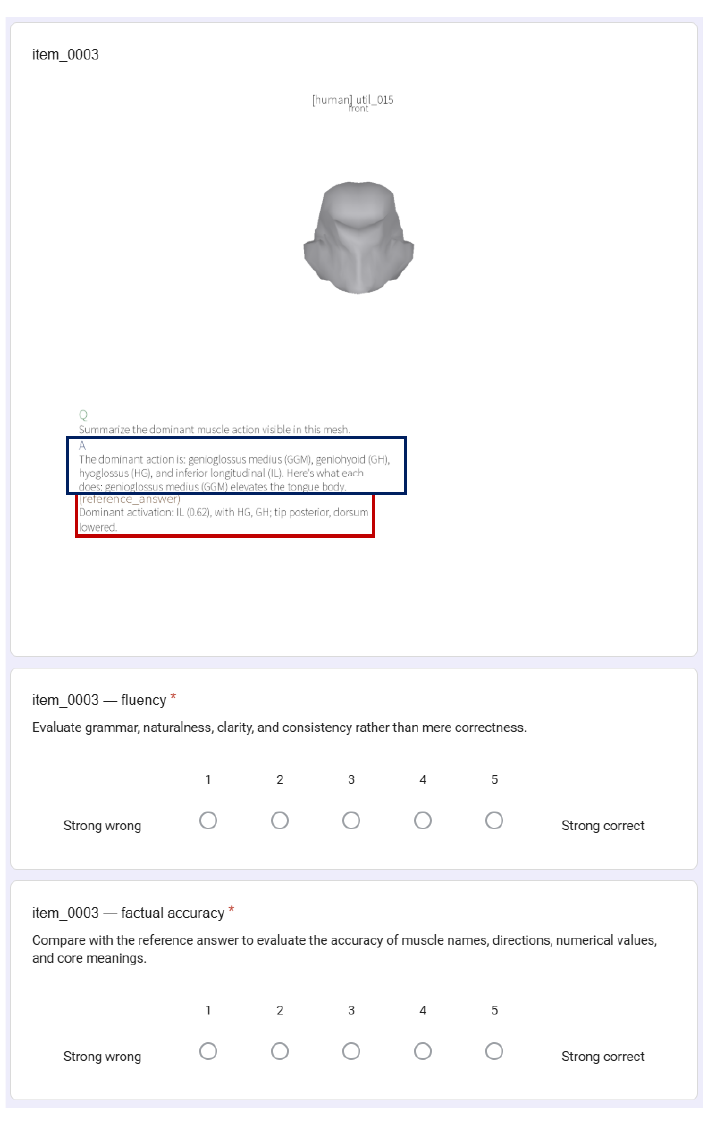}
\caption{Questionnaire interface used for the human evaluation. Each item
shows the mesh, question, anonymized model response, and deterministic
canonical reference. Raters scored Fluency and reference-based Factual
Accuracy on $1$--$5$ scales.}
\label{fig:human_eval_ui}
\end{figure}

\newpage
\noindent\textbf{Human evaluation.}
Three model-blind speech researchers rated responses for Fluency and
reference-based Factual Accuracy on $1$--$5$ scales. The $30$ English prompts
were drawn from question templates held out from training, so their
deterministic references exist but no model saw their question form during
training. Each item displayed the mesh, question, anonymized response, and
deterministic canonical reference.
Question and response order were randomized. One annotator left nine Korean
items unanswered, and these ratings were excluded without imputation. Ratings
were collected with informed consent, and account identifiers were replaced by
anonymous rater codes before analysis. Figure~\ref{fig:human_eval_ui} shows the
questionnaire interface used for the study.

\begin{table}[t]
\centering
\caption{Open-ended human ratings with item-cluster bootstrap $95\%$
confidence intervals over the $30$ evaluated prompts.}
\label{tab:supp_human_results}
\setlength{\tabcolsep}{3pt}
\scriptsize
\resizebox{\columnwidth}{!}{%
\begin{tabular}{lrrr}
\toprule
Model & $n$ & Fluency, mean [95\% CI]
& Factual Accuracy, mean [95\% CI] \\
\midrule
Text-only LLM & 90 & 2.34 [2.09, 2.62] & 1.77 [1.57, 1.99] \\
GPT-5 Pro & 90 & 3.50 [3.09, 3.92] & 2.23 [1.99, 2.50] \\
2D Muscle-Aware + LLM & 90 & 2.86 [2.49, 3.24] & 2.33 [2.08, 2.61] \\
3D Mesh-AE + LLM & 90 & 3.08 [2.72, 3.44] & 2.70 [2.49, 2.92] \\
3D Muscle-Aware + LLM & 90 & 2.72 [2.39, 3.09] & 3.81 [3.50, 4.10] \\
3D Muscle-Aware + LLM, Korean & 81 & 2.86 [2.54, 3.20] & 3.79 [3.52, 4.05] \\
\bottomrule
\end{tabular}%
}
\end{table}

Ordinal Krippendorff's $\alpha$ is $0.560$ for Fluency and $0.622$ for Factual
Accuracy, and the corresponding average-rater agreement is
$\mathrm{ICC}(2,k)=0.816$ and $0.831$. Prompt-level Friedman tests show a
model effect for Fluency with $\chi^2(4)=13.03$ and $p=0.011$ and for Factual
Accuracy with $\chi^2(4)=58.10$ and $p<0.001$. Holm-corrected paired Wilcoxon
tests show that the 3D Muscle-Aware model has higher Factual Accuracy than
each English comparison model, with all adjusted $p\leq0.001$. GPT-5 Pro has
the highest Fluency mean, but its difference from the 3D Muscle-Aware model
is not significant after Holm correction, with $p_{\mathrm{adj}}=0.263$.

\noindent\textbf{Scope and limitations.}
The instantiation uses one Badin anatomy and a fixed topology, and the symmetric model
and midsagittal representation do not fully capture lateral or grooved
articulations. Fixed lips and uncoupled jaw motion limit vocal-tract context and
contribute to compressed $F_2$. Simulator activations are privileged generating
labels rather than uniquely identifiable physiological causes. Consequently,
3DTongueQA supports simulator-grounded articulatory reasoning and future
pronunciation-feedback research, but does not constitute clinical validation or
a replacement for diagnosis or treatment.

\clearpage

\end{document}